%% file: frame.tex
\documentclass{article}

\PassOptionsToPackage{round}{natbib}

\usepackage[final]{neurips_2022}

\usepackage[utf8]{inputenc} 
\usepackage[T1]{fontenc}    
\usepackage[hidelinks]{hyperref}       
\usepackage{url}            
\usepackage{booktabs}       
\usepackage{amsfonts}       
\usepackage{nicefrac}       
\usepackage{microtype}      
\usepackage{xcolor}         

\usepackage{amsmath}
\usepackage{mathrsfs}
\usepackage{amssymb}
\usepackage{amsthm}
\usepackage[ruled]{algorithm2e}
\usepackage{graphicx}
\usepackage{bm}
\usepackage{mathtools}
\usepackage{thmtools,thm-restate}
\usepackage{multirow}
\usepackage{pgf}
\usepackage{subcaption}
\usepackage[american]{babel}
\usepackage{xcolor}

\usepackage{tikz}
\usetikzlibrary{decorations.pathmorphing,decorations.pathreplacing,calligraphy}
\usetikzlibrary{calc,arrows.meta,positioning}

\newtheorem{definition}{Definition}
\newtheorem{theorem}{Theorem}
\newtheorem{corollary}{Corollary}

\newtheorem{proposition}{Proposition}

\newtheorem{innercustomthm}{Theorem}
\newenvironment{customthm}[1]
  {\renewcommand\theinnercustomthm{#1}\innercustomthm}
  {\endinnercustomthm}

\newtheorem{innercustomlemma}{Lemma}
\newenvironment{customlemma}[1]
  {\renewcommand\theinnercustomlemma{#1}\innercustomlemma}
  {\endinnercustomthm}

\newtheorem{innercustomprop}{Proposition}
\newenvironment{customprop}[1]
  {\renewcommand\theinnercustomprop{#1}\innercustomprop}
  {\endinnercustomthm}

\newtheorem{innercustomcor}{Corollary}
\newenvironment{customcor}[1]
  {\renewcommand\theinnercustomcor{#1}\innercustomcor}
  {\endinnercustomthm}

\input{parts/theorems.tex}

\DeclareMathOperator*{\argmax}{\arg\!\max}

\newcommand{\pdel}{\ensuremath{p_{d}}}
\newcommand{\pabl}{\ensuremath{p_{a}}}

\addto\extrasamerican{%
}

\title{Randomized Message-Interception Smoothing: Gray-box Certificates for Graph Neural Networks}

\author{%
Yan Scholten\textsuperscript{\ensuremath{1}}, Jan Schuchardt\textsuperscript{\ensuremath{1}}, Simon Geisler\textsuperscript{\ensuremath{1}},\\ \textbf{Aleksandar Bojchevski\textsuperscript{\ensuremath{2}} \& Stephan Günnemann\textsuperscript{\ensuremath{1}}}\\
\texttt{\{y.scholten, j.schuchardt, s.geisler\}@tum.de}\\ 
\texttt{bojchevski@cispa.de}, \texttt{s.guennemann@tum.de}\\
\textsuperscript{\ensuremath{1}}{Dept. of Computer Science \& Munich Data Science Institute, Technical University of Munich}\\
\textsuperscript{\ensuremath{2}}{CISPA Helmholtz Center for Information Security}
}

\begin{document}

\maketitle

\begin{abstract}
  Randomized smoothing is one of the most promising frameworks 
  for certifying the adversarial robustness of machine learning models, including Graph Neural Networks (GNNs).
  Yet, existing randomized smoothing certificates for GNNs are overly pessimistic since they treat the model as a black box, ignoring the underlying architecture.
  To remedy this, we propose novel gray-box certificates that exploit the message-passing principle of GNNs:
  We randomly intercept messages and carefully analyze the probability 
  that messages from adversarially controlled nodes reach their target nodes.
  Compared to existing certificates, we certify robustness to much stronger adversaries
  that control entire nodes in the graph and can arbitrarily manipulate node features.
  Our certificates provide stronger guarantees for attacks at larger distances, as messages from farther-away nodes are more likely to get intercepted. We demonstrate the effectiveness of our method on various models and datasets. Since our gray-box certificates consider the underlying graph structure, we can significantly improve certifiable robustness by applying graph sparsification.%
  \footnote{Project page: \url{https://www.cs.cit.tum.de/daml/interception-smoothing}}
\end{abstract}

\input{parts/part-1-introduction.tex}
\input{parts/part-2-preliminaries.tex}
\input{parts/part-3-certificate.tex}
\input{parts/part-4-evaluation.tex}
\input{parts/part-5-related-work.tex}

\input{parts/part-6-conclusion.tex}

\begin{ack}
This work has been funded by the German Federal Ministry of Education and Research, the Bavarian State Ministry for Science and the Arts, and the German Research Foundation, grant GU 1409/4-1. The authors of this work take full responsibility for its content.
\end{ack}

\bibliographystyle{plainnat}
\bibliography{references}

\appendix
\clearpage
\input{parts/part-7-appendix.tex}

\end{document}

%% file: parts/theorems.tex
\newcommand{\boundingProposition}[3]{
\begin{customprop}{#1}\ifthenelse{\equal{#2}{}}{}{\label{#2}}\ifthenelse{\equal{#3}{}}{}{\setcounter{proposition}{#3}}
Given target node \(v\) in graph \(G\), and adversarial budget \(\rho\). Let \(E\) denote the event that the prediction \(f_v(\phi(G))\) receives at least one message from perturbed nodes. Then the change in label probability \(|p_{v,y}(G) - p_{v,y}(G')|\) is bounded by the probability \(\Delta = p(E)\) for all classes \(y\in \{1, \ldots, C\}\) and graphs \(G'\) with \(G' \in B_{\rho}(G)\): \(|p_{v,y}(G) - p_{v,y}(G')| \leq \Delta\).
\end{customprop}%
}

\newcommand{\propNodeAblationFirst}[3]{
\begin{customprop}{#1}\ifthenelse{\equal{#2}{}}{}{\label{#2}}\ifthenelse{\equal{#3}{}}{}{\setcounter{proposition}{#3}}
For node feature ablation smoothing only (\(\pdel=0\)), we have \( \Delta = 1-\pabl^\rho \).
\end{customprop}%
}

\newcommand{\messageInterceptionLemma}[3]{
\begin{customlemma}{#1}\ifthenelse{\equal{#2}{}}{}{\label{#2}}\ifthenelse{\equal{#3}{}}{}{\setcounter{lemma}{#3}}
Given a fixed target node \(v\) and perturbed nodes \(\mathbb{B}\) in the graph with \(v\notin \mathbb{B}\). Then \(f_v(\phi(G)) = f_v(\phi(G'))\) for any graph~\(G' \in B_\rho(G)\) if
\[
    \forall w\in\mathbb{B}:
    \left( \forall p\in P^k_{wv}: \exists (i,j)\in p: \phi_1(\bm{A})_{ij} =0 \right) 
    \vee
    \left( \phi_2(\bm{x}_w) = \bm{t} \right)
\]
\end{customlemma}%
}

\newcommand{\boundingTheorem}[3]{
\begin{customthm}{#1}\ifthenelse{\equal{#2}{}}{}{\label{#2}}\ifthenelse{\equal{#3}{}}{}{\setcounter{theorem}{#3}}
The worst-case change in label probability \(|p_{v,y}(G) - p_{v,y}(G')|\) is bounded by
\[
    \Delta = \max_{||\bm{\rho}_v||_0\leq \rho} p\left(E({\bm{\rho}_v})\right) 
\]
for all classes \(y\in \{1, \ldots, C\}\) and any graph \(G' \in B_{\rho}(G)\).
\end{customthm}%
}

\newcommand{\corMulticlassCert}[3]{
\begin{customcor}{#1}[Multi-class certificate]\ifthenelse{\equal{#2}{}}{}{\label{#2}}\ifthenelse{\equal{#3}{}}{}{\setcounter{corollary}{#3}}
  Given \(\Delta\) as defined in \autoref{proposition:boundingProposition}.
  Then we can certify the robustness \(g_v(G) = g_v(G')\) for any graph \(G' \in B_{\rho}(G)\) if
  \[ p_{v,y^*}(G) - \Delta > \max_{\tilde{y}\neq y^*} p_{v,\tilde{y}}(G) + \Delta \]
  where \(y^* \triangleq g_v(G)\) denotes the majority class, and \(\tilde{y}\) the follow-up (second best) class.
\end{customcor}%
}

\newcommand{\singleSourceMB}[3]{
\begin{customthm}{#1}[Single Source Multiplicative Bound]\ifthenelse{\equal{#2}{}}{}{\label{#2}}\ifthenelse{\equal{#3}{}}{}{\setcounter{theorem}{#3}}
  Given target node \(v\) and source node \(w\neq v\) in the receptive field of a k-layer message-passing GNN \(f\) with respect to \(v\). Let \(P_{wv}^k\) denote all simple paths from~\(w\) to~\(v\) of length at most \(k\) in graph \(G\). Then \(\Delta_w\leq\overline{\Delta}_w\) for:
  \[ 
    \overline{\Delta}_w \triangleq \left[ 1-\prod_{q\in{P_{wv}^k}} \left( 1- (1-\pdel)^{|q|} \right) \right](1-\pabl) 
  \]
  where \(|q|\) denotes the number of edges on the simple path \(q\in P_{wv}^k\) from \(w\) to \(v\).
\end{customthm}%
}

\newcommand{\theoremMultiplicativeBound}[3]{
\begin{customthm}{#1}[Generalized multiplicative bound]\ifthenelse{\equal{#2}{}}{}{\label{#2}}\ifthenelse{\equal{#3}{}}{}{\setcounter{theorem}{#3}}
  Assume an adversarial budget of \(\rho\) nodes and
  let \(\Delta_1, \ldots, \Delta_\rho\) denote the \(\rho\) largest \(\Delta_i\) for nodes \(i\) in the receptive field. Then we have \(\Delta \leq \overline{\Delta}_M\) for
  \[ \overline{\Delta}_M \triangleq 1-\prod_{i=1}^{\rho} (1-\Delta_i) \] 
\end{customthm}%
}

\newcommand{\corollaryPracticalCerts}[3]{
\begin{customcor}{#1}\ifthenelse{\equal{#2}{}}{}{\label{#2}}\ifthenelse{\equal{#3}{}}{}{\setcounter{corollary}{#3}}
  We guarantee \(g_v(G) = g_v(G')\) with probability of at least \(1-\alpha\) for any \(G' \in B_{\rho}(G)\) if
  \( 
    \underline{p_{v,y^*}(G)} - \overline{\Delta} > \overline{p_{v,\tilde{y}}(G)} + \overline{\Delta}
  \), where \(y^* \) denotes the majority class, and \(\tilde{y}\) the follow-up~class.
\end{customcor}%
}

\newcommand{\propNodeAblationSecond}[3]{
\begin{customprop}{#1}\ifthenelse{\equal{#2}{}}{}{\label{#2}}\ifthenelse{\equal{#3}{}}{}{\setcounter{proposition}{#3}}
  Given fixed \(\pabl>0\) and \(\pdel=0\), it is impossible to certify a radius \(\rho\) if \(\pabl \leq \sqrt[\rho]{0.5}\).
\end{customprop}%
}

%% file: parts/part-1-introduction.tex
\section{Introduction}

The core principle behind the majority of Graph Neural Networks (GNNs) is message passing -- the representation of a node is (recursively) computed based on the representations of its neighbors \citep{gilmer2017neural}. This allows for information to propagate across the graph, e.g.\ in a k-layer GNN the prediction for a node depends on the messages received from its k-hop neighborhood. With such models, if an adversary controls a few nodes in the graph, they can manipulate node features to craft adversarial messages that in turn change the prediction for a target node.

Such feature-based adversarial attacks are becoming significantly stronger in recent years and pose a realistic threat \citep{ma2020towards,zou2021tdgia}: Adversaries may arbitrarily manipulate features of entire nodes in their control, for example in social networks, public knowledge graphs and graphs in the financial and medical domains. Detecting such adversarial perturbations is a difficult unsolved task even beyond graphs \citep{carlini2017adversarial}, meaning such attacks may go~unnoticed.

\begin{figure}[t]
    \centering
    \input{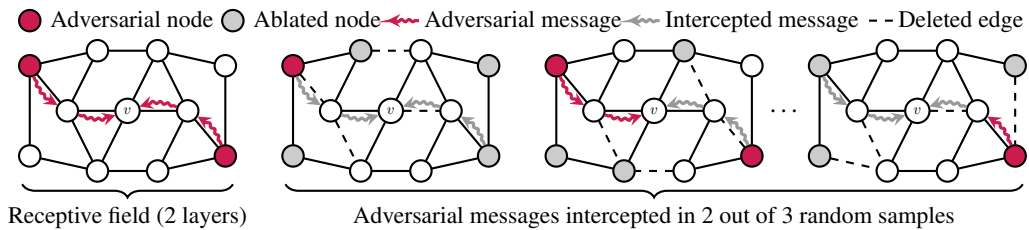}
    \caption{
		Randomized message-interception smoothing: We model adversaries that can arbitrarily manipulate features of multiple nodes in their control (red) to alter the predictions for a target node~\(v\). We intercept messages (gray) by randomly deleting edges and/or ablating (mask) all features of entire nodes. Our certificates are based on the majority vote under this randomized message~interception.
	}
    \label{fig:firstfig}
\end{figure}

How can we limit the influence of such adversarial attacks? We introduce a simple but powerful~idea: \emph{intercept} adversarial messages. Specifically, we propose message-interception smoothing where we randomly delete edges and/or randomly ablate (mask) nodes, and analyze the probability that messages from adversarially controlled nodes reach the target nodes. By transforming any message-passing GNN into a smoothed GNN, where the prediction is the majority vote under this randomized message interception, we can provide robustness certificates (see \autoref{fig:firstfig}).

Experimentally we obtain significantly better robustness guarantees compared  to previous (smoothing) certificates for GNNs (compare Section \ref{sec:experiments}). This improvement stems from the fact that our certificates take the underlying architecture of the classifier into account. Unlike previous randomized smoothing certificates which treat the GNN as a black-box, our certificates are \emph{gray-box}. By making the certificate message-passing aware we partially open the black-box and obtain stronger guarantees.

Our approach is also in contrast to white-box certificates that apply only to very specific models. For example, \citet{zuegner2019certifiable} only certify the GCN model \citep{kipf2017semi}. While newly introduced GNNs require such certificates to be derived from scratch, our approach is model-agnostic and flexible enough to accommodate the large family of message-passing GNNs.

We evaluate our certificates on node classification datasets and analyze the robustness of existing GNN architectures. By applying simple graph sparsification we further increase the certifiable robustness while retaining high accuracy, as sparsification reduces the number of messages to intercept. In stark contrast to previous probabilistic smoothing-based certificates for GNNs, our certificates require only a few Monte-Carlo samples and are more efficient: For example, we can compute certificates on Cora-ML in just 17 seconds and certify robustness against much stronger adversaries than previous smoothing-based certificates \citep{bojchevski2020efficient} that take up to 25~minutes.

In short, our main contributions~are: 
\begin{itemize}
    \item The first gray-box smoothing-based certificates for GNNs that exploit the underlying \emph{message-passing} principle for stronger guarantees.
	\item Novel randomized smoothing certificates for strong threat models where adversaries can arbitrarily manipulate features of multiple nodes in their control.
\end{itemize}

%% file: parts/part-2-preliminaries.tex
\section{Preliminaries and Background} 

\textbf{Threat model.}
We develop certificates for feature perturbations given \textit{evasion} threat models. Specifically, we model adversaries that attack GNNs by entirely perturbing attributes of a few \(\rho\) nodes in the graph at inference. Given an attributed graph \(G=(\bm{A},\bm{X})\in\mathbb{G}\) encoded via adjacency matrix \(\bm{A} \in \{0,1\}^{n\times n}\) and feature matrix \(\bm{X}\in \mathbb{R}^{n\times d}\) with \(n\) nodes and \(d\) features, we formally define the threat model of feature perturbations as a ball centered at a given graph \(G = (\bm{A}, \bm{X})\):
\[ 
  B_{\rho}(G) \triangleq \{ G' =  (\bm{A}', \bm{X}') \mid \bm{A} = \bm{A'}, \delta(G, G') \leq \rho \}
\]
where
\( 
  \delta(G, G') \triangleq \sum_{v=1}^{n} \bm{1}_{\bm{x}_{v}\neq \bm{x}'_{v}}
\)
denotes the number of nodes whose features differ in at least one dimension when comparing the clean graph \(G\) and the perturbed graph~\(G'\). Intuitively, this means adversaries control up to \(\rho\) nodes in the graph and can arbitrarily manipulate node features.

\textbf{Graph neural networks.}
We design robustness certificates for GNNs that instantiate the so-called message-passing framework \citep{gilmer2017neural}. The message-passing framework describes a large family of GNN architectures that are based on the local aggregation of information from neighboring nodes in the graph. To compute a new representation~\(\bm{h}_{v}^{(\ell)}\) of node~\(v\), each message-passing layer \(\Psi^{(\ell)}\) transforms and aggregates the representations \(\bm{h}_{v}^{(\ell-1)}\) and \(\bm{h}_{u}^{(\ell-1)}\) of all nodes~\(u\) in the local neighborhood \(\mathcal{N}(v) \triangleq \{ u \mid \bm{A}_{uv} = 1\}\) of node~\(v\). 

We can formally describe a message-passing layer as follows:
\(
  \bm{h}_v^{(\ell)} \triangleq \Psi_{u\in\mathcal{N}(v)\cup\{v\}}^{(\ell)}
    \left( \bm{h}_{v}^{(\ell-1)},\bm{h}_{u}^{(\ell-1)}\right)
\).
For node classification, message-passing GNNs with \(k\) GNN-layers can be described as parametrized functions \(f : \mathbb{G} \rightarrow \{1, \ldots, C\}^n\) that assign each node~\(v\) in graph \(G\) class
\(
  f_v(G) \triangleq \argmax_c \bm{h}_{v,c}^{(k)}
\), where~\(\bm{h}_v^{(0)} \triangleq \bm{x}_{v} \in \mathbb{R}^d\) denotes the input and \(\bm{h}_v^{(k)}\in\mathbb{R}^C\) the final representation of node~\(v\).

\textbf{Randomized smoothing.}
Our robustness certificates for GNNs build upon the randomized smoothing framework \citep{cohen2019certified,lecuyer2019certified}: Given any base classifier \(f\), for example a message-passing GNN, we can build a ``smoothed'' classifier~\(g\) that  classifies randomly perturbed input samples, and then takes the ``majority vote'' among all predictions. The goal is to construct a smoothed classifier that behaves similar to \(f\) (for example in terms of accuracy) and for which we can prove (probabilistic) robustness certificates. 

Randomized ablation \citep{levine2020robustness} is a smoothing-based certificate that ``ablates'' the~input: Unlike in randomized smoothing where the input is randomly perturbed (e.g. by adding Gaussian noise to images), in randomized ablation the input is randomly masked, for example by replacing parts of the input with a special ablation token that ``hides'' the original information. If the perturbed input is masked for the majority of predictions, we can issue certificates for the smoothed classifier~\(g\).

%% file: parts/part-3-certificate.tex
\section{Randomized Message-Interception Smoothing for Graph Neural Networks}

The main idea of our gray-box smoothing certificates is to intercept messages from perturbed nodes by (1) deleting edges to disconnect nodes, and/or (2) ablating nodes to mask their features (cf.~\autoref{fig:firstfig}).

To implement this we introduce two independent smoothing distributions \(\phi_1(\bm{A})\) and \(\phi_2(\bm{X})\) that randomly apply these changes to the input graph: The first smoothing distribution \(\phi_1(\bm{A})\) randomly deletes edges in the adjacency matrix (\(1\rightarrow 0\)) with probability \(\pdel\). The second smoothing distribution \(\phi_2(\bm{X})\) randomly ablates all features of nodes with probability \(\pabl\) by replacing their feature representations with a fixed representation token \(\bm{t}\in\mathbb{R}^d\) for ablated nodes. The ablation representation \(\bm{t}\) is a trainable parameter of our smoothed classifier and can be optimized during training. Introducing two independent smoothing distributions is important since our base classifiers~\(f\) are GNNs, which behave differently under structural changes in the graph than to feature ablation of nodes in practice.

We use this message-interception smoothing distribution \(\phi(G) \triangleq (\phi_1(\bm{A}), \phi_2(\bm{X}))\) to randomly sample and then classify different graphs with a message-passing GNN \(f\). Finally, our smoothed classifier \(g\) takes the majority vote among the predictions of \(f\) for the sampled graphs \(\phi(G)\). We formally describe our smoothed classifier \(g\) as follows: 
\[ g_{v}(G) \triangleq \argmax_{y\in\{1,\ldots,C\}} p_{v,y}(G) \quad\quad p_{v,y}(G) \triangleq p(f_v(\phi(G))=y)\]
where \(p_{v,y}(G)\) denotes the probability that the base GNN \(f\) classifies node \(v\) in graph \(G\) as class~\(y\) under the smoothing distribution \(\phi(G) = (\phi_1(\bm{A}), \phi_2(\bm{X})) \).

\section{Provable Gray-box Robustness Certificates for Graph Neural Networks}\label{sec:certificates}
We derive provable certificates for the smoothed classifier \(g\). To this end, we develop a condition that guarantees \(g_v(G) = g_v(G')\) for any graph~\(G' \in B_\rho(G)\): We make the worst-case assumption that adversaries alter the prediction for a target node whenever it receives at least one message from perturbed nodes. Let \(E\) denote the event that at least one message from perturbed nodes reaches a target node \(v\). Then the probability \(\Delta \triangleq p(E)\) quantifies how much probability mass of the distribution \(p_{v,y}(G)\) over classes \(y\) is controlled by the worst-case~adversary:%
\boundingProposition{1}{proposition:boundingProposition}{}
{\textit{Proof sketch} (Proof~in~\autoref{appendix:certificateProofs}). Whenever we intercept all adversarial messages, adversaries cannot alter the~prediction. Thus \(|p_{v,y}(G) - p_{v,y}(G')|\) is bounded by \(\Delta\).\hfill\(\square\)}%

Note that we derive an upper bound on \(\Delta\) in \autoref{sec:practical}.

We first consider the special case of node ablation smoothing, discuss its relation to randomized ablation for image classifiers \citep{levine2020robustness}, and then we derive our provably stronger guarantees for the general case of message-interception smoothing.

\textbf{Special case of node ablation smoothing.}
For the special case of node feature ablation smoothing only (\(\pdel=0\)),
we can directly determine the probability \(\Delta\) (Proof in \autoref{appendix:nodeAblation}):
\propNodeAblationFirst{2}{propNodeAblationFirst}{2}
In this special case, our certificates for GNNs are theoretically related to the randomized ablation certificates for image classifiers \citep{levine2020robustness}. We could apply their smoothing distribution to GNNs by randomly ablating features of entire nodes, instead of pixels in an image. However, their approach is specifically designed for image classifiers and comes with serious shortcomings when applied to GNNs. Notably, our robustness cetificates are provably tighter and experimentally stronger even in this special case without edge deletion smoothing (\(\pdel=0\)): Given that \(\Delta^L\) denotes the bounding constant as defined by \citet{levine2020robustness}, we show \(\Delta < \Delta^L\) in \autoref{appendix:nodeAblation}. We carefully discuss such differences with more technical details in \autoref{appendix:comparisonLevine}. Most importantly, their certificate applied to GNNs ignores the underlying graph structure.

\textbf{General case of message-interception smoothing.}
In contrast, our message-interception certificates are specifically designed for \emph{graph-structured} data, message-passing aware, and consider the interception of messages via edge deletion as follows:

Consider a fixed target node \(v\) in the graph. The formal condition for intercepting messages from a fixed target node \(v\) to itself is \(\phi_2(\bm{x}_v) = \bm{t}\),
since we only intercept messages from the target node to the target node itself if we ablate its features. To model the interception of messages from perturbed nodes \(\mathbb{B}\) other than the target node, we take~the graph structure~\(\bm{A}\) into account: We consider all simple paths \(P^k_{wv} = \{ (e_1, \ldots, e_i) \mid i \leq k \}\) from perturbed nodes \(w\in\mathbb{B}\) to target node \(v\) of length at most \(k\) (where \(k\) is the number of GNN layers).%
\footnote{We consider simple paths (all nodes appear only once), since we only receive perturbed messages via more complex paths iff we receive perturbed messages via the simple part of the complex path.}
Intuitively, if any edge~\(e\)~on~path \(p\in P^k_{wv}\) is deleted, or the features of \(w\) are ablated, messages via path \(p\) get intercepted. If all messages from perturbed nodes get intercepted, adversaries cannot alter the prediction for the target node (Proof~in~\autoref{appendix:certificateProofs}):
\messageInterceptionLemma{1}{messageInterceptionLemma}{1}
Since k-layer message-passing GNNs aggregate information over local neighborhoods, only features of nodes in the \emph{receptive field} affect the prediction for a target node (only via paths with a length of at most~\(k\) to \(v\)). For any perturbed node \(w\in\mathbb{B}\) outside of the receptive field we have \(P^k_{wv}=\emptyset\) and the message-interception condition of \autoref{messageInterceptionLemma} is always fulfilled.

In practice, however, we do not know which nodes in the graph are controlled by the adversary. To account for this, we assume adversaries control nodes indicated by \(\bm{\rho}_v \in \{0,1\}^n\) that maximize the probability of the event \( E(\bm{\rho}_v) \) that target node \(v\) receives perturbed messages:
\boundingTheorem{1}{boundingTheorem}{1}
Proof in \autoref{appendix:certificateProofs}. Finally, we provide conservative robustness certificates for the smoothed classifier \(g\) by exploiting that perturbed nodes are disconnected and/or ablated and cannot send messages for the majority of predictions:
\corMulticlassCert{1}{corMulticlassCert}{1}
Proof in \autoref{appendix:certificateProofs}. We also provide a certificate for binary node classification in \autoref{appendix:certificateProofs}.

\section{Practical Interception Smoothing Certificates}\label{sec:practical}

Message-interception certificates constitute two challenges in practice: (1) computing the bounding constant \(\Delta\) for arbitrary graphs, and (2) computing the label probabilities \(p_{v,y^*}(G)\) and \(p_{v,\tilde{y}}(G)\). We address the first problem by providing upper bounds on~\(\Delta\) (i.e. lower bounds on the certifiable robustness). For the second problem we follow existing literature and estimate the smoothed~classifier.

\textbf{Lower bound on certifiable robustness.}
Computing \(\Delta\) of \autoref{boundingTheorem} poses two problems: First, finding the worst-case nodes in arbitrary graphs involves a challenging optimization over the powerset of nodes in the receptive field. Second, computing the probability \( p(E(\bm{\rho}_v))\) to receive perturbed messages is challenging even for fixed \(\bm{\rho}_v\), since in general, it involves evaluating the inclusion-exclusion principle (\autoref{appendix:ieprinciple}). We can compute \(\Delta\) exactly only for special cases such as small or tree-structured receptive fields~(\autoref{appendix:trees}). Notwithstanding the challenges, we provide practical upper bounds on \(\Delta\). Instead of assuming a fixed \(\bm{\rho}_v\), we solve both problems regarding \(\Delta\) at once and directly bound the maximum over all possible \(\bm{\rho}_v\) by assuming \textit{independence} between paths. Due to \autoref{corMulticlassCert}, any upper bound on \(\Delta\) result in lower bounds on the certifiable~robustness.

\begin{figure}[t]
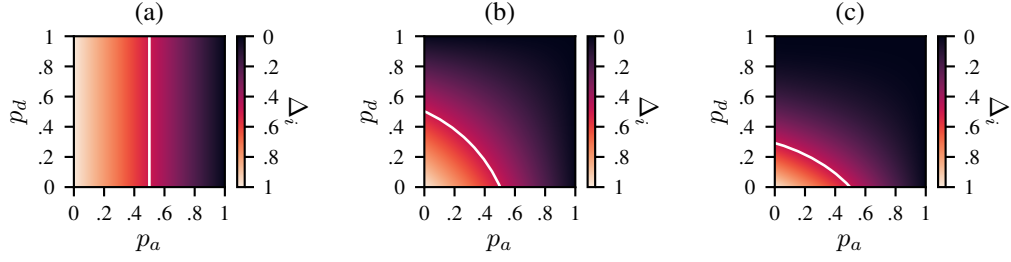

  \centering
  \begin{minipage}{0.333\textwidth}
      \centering
      \input{figures/fig17.pgf}
  \end{minipage}\hfill%
  \begin{minipage}{0.333\textwidth}
      \centering
      \input{figures/fig16.pgf}
  \end{minipage}\hfill%
  \begin{minipage}{0.333\textwidth}
      \centering
      \input{figures/fig15.pgf}
  \end{minipage}
  \caption{
    Single source bounding constant \(\Delta_i\) for different
    edge deletion probabilities \(\pdel\) and node feature ablation probabilities \(\pabl\). White isolines indicate \(\Delta_i=0.5\) and separate the 
    theoretically certifiable region (\(\Delta_i < 0.5\)) from the
    uncertifiable region (\(\Delta_i \geq 0.5\)).
    (a) For the target node, \(\pdel\) does not affect \(\Delta_i\). 
    (b) Direct neighbor of target node, single edge.
    (c) Second-hop neighbor, single path (two edges).
    (a-c) More distant nodes have larger theoretically certifiable regions.
  }\label{fig:SingleSourceDelta}
\end{figure}

We first derive an upper bound on \(\Delta\) for a single perturbed node, and then generalize to multiple nodes. Let \(E_w\) denote the event that the target node \(v\) receives messages from node \(w\), and \( \Delta_w \triangleq p(E_w)\). Note in the special case of the target node \(v=w\) we just have \(\Delta_w = 1-\pabl\), since the features \(\bm{x}_v\) of the target node \(v\) are used for the prediction independent of any~edges. For any \(w\neq v\) in the receptive field we can derive the following upper bound for single sources (Proof in \autoref{appendix:boundProofs}):
\singleSourceMB{2}{singleSourceMB}{2}
We visualize \(\Delta_w\) for different \(\pdel\) and \(\pabl\) in \autoref{fig:SingleSourceDelta}. The upper bound for single sources is tight 
for one- and two-layer GNNs (\(\Delta=\overline{\Delta}_w\)), since then all paths from a single source to the target node are independent (\autoref{appendix:boundProofs}).
The single source multiplicative bound on \(\Delta_w\) can only be used to certify a radius of \(\rho=1\). For multiple nodes (\(\rho > 1\)), we generalize \autoref{singleSourceMB} as follows:%
\theoremMultiplicativeBound{3}{theoremMultiplicativeBound}{3}
Proof in \autoref{appendix:boundProofs}. Notably, the multiplicative bound is tighter than a union bound. We specifically address the approximation error in detail in \autoref{appendix:approximationError}.

\paragraph{Estimating the smoothed classifier in practice.}\label{sec:finalPracticalCerts}
Computing the probabilities \(p_{v,y^*}(G)\) and \(p_{v,\tilde{y}}(G)\) exactly is challenging in practice. We instead estimate them similar to previous work by drawing Monte-Carlo samples from \(\phi\) \citep{cohen2019certified,levine2020robustness,bojchevski2020efficient}. We first identify the majority class \(y^*\) and follow-up class \(\tilde{y}\) using a few samples. We then draw more samples to estimate a lower bound \(\underline{p_{v,y^*}(G)}\) on \(p_{v,y^*}(G)\) and an upper bound \(\overline{p_{v,\tilde{y}}(G)}\) on~\(p_{v,\tilde{y}}(G)\). We use the Clopper-Pearson Bernoulli confidence interval and apply Bonferroni correction to ensure that the bounds hold simultaneously with significance level~\(\alpha\) (with probability of at least \(1-\alpha\)). Moreover, our smoothed classifier abstains from predicting if \(\underline{p_{v,y^*}(G)} \leq \overline{p_{v,\tilde{y}}(G)}\), meaning if the estimated probabilities are too similar. We experimentally analyze abstained predictions in~\autoref{appendix:fullresults}.

\textbf{Practical robustness certificates.}
Finally, our robustness certificates also hold when bounding \(\Delta\) and the label probabilities as the following Corollary shows (Proof in \autoref{appendix:certificateProofs}):
\corollaryPracticalCerts{2}{corollaryPracticalCerts}{2}

\section{Discussion}\label{sec:discussion}

Our certificates require knowledge about the graph structure \(\bm{A}\) and can only account for structure perturbations if the perturbed adjacency matrix \(\bm{A}'\) is known. While adversarial edge deletion potentially increases robustness (due to less messages to intercept), adversaries could arbitrarily increase the number of messages via edge insertion. Moreover, the number of simple paths in the graph can be huge. We argue, however, that (1) graphs are usually sparse, (2) the number of paths can be reduced via sparsification, and (3) we have to compute paths only once for each~graph.

\textbf{Limitations of ablation certificates.} Since the probability to receive messages from perturbed nodes increases the more nodes are adversarial, \(\Delta\) is monotonously increasing in \(\rho\). Thus, the certifiable radius is bounded independent of the label probabilities (uncertifiable region for \(\Delta\geq 0.5\) due to \autoref{corMulticlassCert}). This bound depends on the graph structure and changes for each target node, but in the case of node feature ablation smoothing we can directly determine the bound (Proof in~\autoref{appendix:npandablation}):%
\propNodeAblationSecond{3}{propNodeAblationSecond}{3}
This bound is only determined by the parameters of the smoothing distribution (\(\pdel,\pabl\)) and does not depend on the base GNN \(f\). The existence of an upper bound is in stark contrast to certificates whose largest certifiable radius depends on the inverse Gaussian CDF of the label probabilities \citep{cohen2019certified}. Such certificates are theoretically tighter than ablation certificates: For example, if the base classifier \(f\) classifies all samples from \(\phi\) the same (\(p_{y^*}=1\)), they would certify a radius of~\(\infty\), whereas the radius of ablation-based certificates is bounded. We leave the development of even stronger gray-box certificates for GNNs to future~work.

\textbf{Limitations of probabilistic certificates.} Our certificates are probabilistic and hold with significance level \(\alpha\).  Notably, our method still yields strong guarantees for significantly smaller confidence levels (we show additional experiments for varying \(\alpha\) in \autoref{appendix:alpha}). We found that \(\alpha\) has just a minor effect on the certificate strength, since increasing it cannot increase the largest certifiable radius, which is theoretically bounded. Recent works also ``derandomize'' probabilistic certificates, that is they compute the label probabilities exactly \citep{levine2020Derandomized,levine2021improved}. In \autoref{appendix:derandomization} we propose the first derandomization technique that leverages message-passing~structures. We believe future work can build upon it towards even more efficient derandomization schemes.

\textbf{Threat model extensions.} Notably, edge-deletion smoothing (\(\pdel>0\)) also yields guarantees for adversarial node insertion and deletion,
as disconnected nodes cannot alter the prediction.%
\footnote{We cannot certify node insertion/deletion with feature ablation smoothing, since e.g.\ new nodes affect the smoothed classifier independent of whether features are ablated or not (unless we delete nodes entirely).}
As discussed above, we can only evaluate such certificates with structural information, that is how inserted/deleted nodes are connected to target nodes: Given clean graphs (as in our evaluation), we know which nodes adversaries \textit{could delete}. Given perturbed graphs, we know which nodes \textit{could have been~inserted}. Note that although we can technically extend our method to certify adversarial edge deletion, we focus on the novel problem of arbitrary feature manipulations of entire nodes since there are already certificates against edge-modification attacks \citep{bojchevski2020efficient}. 

%% file: parts/part-4-evaluation.tex
\section{Experimental Evaluation}\label{sec:experiments} 

We evaluate our certificates for different GNN architectures trained on node classification datasets. Our certificates work in standard transductive learning settings used throughout the literature and we report such results in \autoref{appendix:fullresults}. However, combining transductive learning with an evasion threat model comes with serious shortcomings for the evaluation of certificates, since no separate test data is available. For example, we can usually achieve high accuracy by overfitting a Multi-Layer Perceptron (MLP) to labels predicted by GNNs during training. MLPs do not propagate information through the graph at test time and are robust to adversarial messages. Instead, we evaluate our certificates in semi-supervised \textit{inductive} learning settings with hold-out test nodes:

\textbf{Experimental setup.}
As labelled nodes, we draw 20 nodes per class for training and validation, and 10\% of the nodes for testing. We use the labelled training nodes and all remaining unlabeled nodes as training graph, and successively insert (hold-out) validation and test nodes. We train on the training graph, optimize hyperparameters against validation nodes, assume adversaries control nodes at test time, and compute certificates for all test nodes. We also delete edges and ablate node features during training (\autoref{appendix:hyperparameters}). We use \(n_0=1{,}000\) samples for estimating the majority class, \(n_1=3{,}000\) samples for certification, and set \(\alpha=0.01\). We conduct five experiments for random splits and model initializations, and report averaged results including standard deviation (shaded areas in the plots). When comparing settings (e.g.\ architectures), we run \(1{,}000\) experiments for each setting and draw deletion and ablation probabilities from \([0,1]\) for each experiment (sampling separately for training and inference). Then, we compute dominating points on the Pareto front for each setting. For brevity, we only show points with clean accuracy of at most \(5\%\) below the maximally achieved~performance.

\textbf{Datasets and models.}
We train our models on citation datasets: Cora-ML \citep{bojchevski2018deep,mccallum2000automating}  with 2,810 nodes, 7,981 edges and 7 classes; Citeseer \citep{sen2008collective} with 2,110 nodes, 3,668 edges and 6 classes; and PubMed \citep{namata2012query} with 19,717 nodes, 44,324 edges and 3 classes. We implement smoothed classifiers for four architectures with two message-passing layers: Graph convolutional networks (GCN) \citep{kipf2017semi}, graph attention networks (GAT and GATv2) \citep{velivckovic2018graph,brody2021how}, and soft medoid aggregation networks (SMA) \citep{geisler2020reliable}. More details in~\autoref{appendix:hyperparameters}. We also compute certificates for the larger graph ogbn-arxiv \citep{hu2020ogb} in \autoref{appendix:arxiv}.

\textbf{Evaluation metrics.}
We report the classification accuracy of the smoothed classifier on the test set (\textit{clean accuracy}), and the \textit{certified ratio}, that is the number of test nodes whose predictions are certifiable robust for a given radius. Since all nodes have different receptive field sizes, we also divide the certifiable radius by the receptive field size. The resulting \textit{normalized} robustness better reflects how much percentage of the ``attack surface'' (that is the number of nodes the adversary could attack) can be certified. Moreover, we report the area under this (normalized) certified ratio curve (\textit{AUCRC}). For completeness, we also report the \textit{certified accuracy} in \autoref{appendix:fullresults}, that is the number of test nodes that are correctly classified (without abstaining) \textit{and} certifiable robust for a given radius.

\begin{figure}[h]
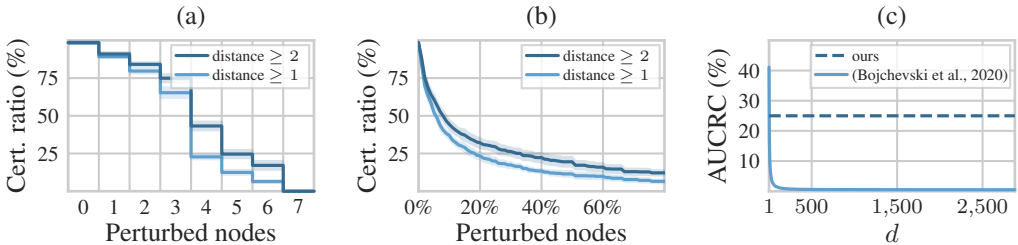

  \centering
  \begin{minipage}{0.333\textwidth}
    \centering
    \input{figures/fig28.pgf}
  \end{minipage}\hfill%
  \begin{minipage}{0.333\textwidth}
    \centering
    \input{figures/fig29.pgf}
  \end{minipage}\hfill%
  \begin{minipage}{0.333\textwidth}
    \centering
    \input{figures/fig35.pgf}
  \end{minipage}
  \caption{Smoothed GAT on Cora-ML:
  (a)~Robustness at different distances to target nodes (\(\pdel\)=\(0.31\), \(\pabl\)=\(0.794\), with skip, ACC=\(0.79\)).
  (b)~Robustness normalized by receptive field size (``attack surface''). 
  (c)~Na\"ive baseline comparison 
  (base certificate \citep{bojchevski2020efficient}, \(10^5\) samples, \(\alpha\)=\(0.01\)). 
  }\label{fig:skipConnection}
\end{figure}

\textbf{Message-interception smoothing.} In \autoref{fig:skipConnection} (a,b) we demonstrate our certificates for specific edge deletion probabilities \(\pdel\) and node feature ablation probabilities \(\pabl\). By making our certificates message-passing aware, we can (1) certify robustness against arbitrary feature perturbations of entire nodes, (2) analyze robustness locally in the receptive fields by incorporating the ``attack surface'', and (3) provide stronger guarantees for attacks against nodes at larger distances to target nodes.

\textbf{First certificate for stronger adversaries.} Experimentally we obtain significantly better robustness guarantees compared to previous (smoothing-based) certificates for Graph Neural Networks. Specifically, existing certificates for GNNs only certify perturbations to a few attributes \(\tilde{\rho}\) in the entire graph. Our certificates are novel as they provide guarantees for much stronger adversaries that can arbitrarily manipulate features of a multiple nodes in the~graph. To compare these two approaches, consider a na\"ive baseline that certifies \(\rho=\tilde{\rho}/d\) nodes, where \(d\) is the number of attributes per node.%
\footnote{We are the first to certify such strong adversaries. Thus no baselines exist so far and we compare our method against existing certificates for GNNs using the na\"ive baseline we propose~above.}
If each node in the graph had just a single feature, the number of certifiable nodes \(\rho\) is high. As the number of features \(d\) per node increases, however, the baseline dramatically deteriorates. In contrast, our certificates are entirely independent of the dimension \(d\) and hold regardless of how high-dimensional the underlying node data might be. We demonstrate this comparison in \autoref{fig:skipConnection}~(c) for the first smoothing-based certificate for GNNs \citep{bojchevski2020efficient}, assuming attribute deletions against second-hop nodes (\(p_{+}\)=\(0\), \(p_{-}\)=\(0.6\)). However, the superiority of our certificate regarding robustness against all features of entire nodes holds for any other GNN certificate proposed so far.

\begin{figure}[h!]
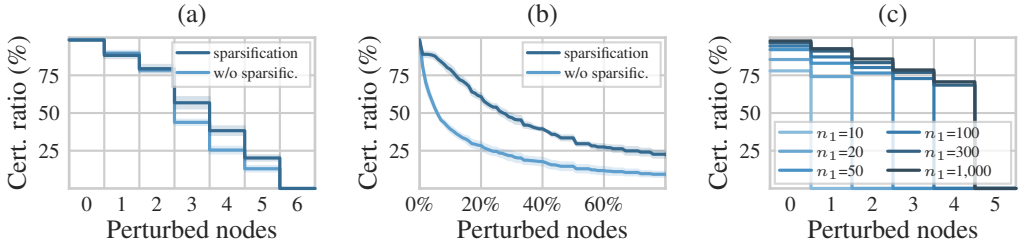

  \centering
  \begin{minipage}[t]{0.333\textwidth}
    \centering%
    \input{figures/fig32.pgf}%
  \end{minipage}\hfill%
  \begin{minipage}[t]{0.333\textwidth}
    \centering%
    \input{figures/fig33.pgf}%
  \end{minipage}\hfill%
  \begin{minipage}[t]{0.333\textwidth}
    \centering%
    \input{figures/fig61.pgf}%
  \end{minipage}
  \caption{
  (a,b)~Sparsification significantly improves certifiable robustness of our gray-box certificates to second-hop attacks since sparsification reduces (a)~messages to intercept, and (b)~receptive field sizes and thus the ``attack surface'' (Smoothed GAT, Cora-ML, \(\pdel=0.31\), \(\pabl=0.71\), with skip-connection, ACC \(=0.8\)). (c) Our certificate with largest certifiable radius of 4 with varying samples for certification~(Smoothed GAT, Cora-ML, \(\pdel=0\), \(\pabl=0.85\)). Our certificates are more sample efficient than existing smoothing-based certificates for GNNs.}\label{fig:archAndSparsification}
\end{figure}

\textbf{Stronger certificates for sparser graphs.}
Notably, our gray-box certificates incorporate graph structure and become stronger for sparser graphs. This is in contrast to black-box certificates that ignore the underlying message-passing principles of GNNs. We demonstrate this by applying graph sparsification, which significantly improves robustness while retaining high clean accuracy: First, sparsification reduces the number of paths in the graph and thus reduces the number of messages to intercept. Second, sparsification reduces the number of nodes in the receptive fields and thus the ``attack surface'', that is the number of nodes that send messages.  In \autoref{fig:archAndSparsification}~(a,b) we apply GDC preprocessing \citep{gasteiger2019diffusion} to the Cora-ML graph at test time. GDC preprocessing yields directed graphs and reduces the number of edges in the graph from \(15{,}962\) to \(14{,}606\) (we set the sparsification
threshold of GDC to \(\epsilon=0.022\) and ignore resulting edge attributes). Interestingly, evaluating the model on the sparsified graph yields significantly higher certifiable robustness, although both approaches show high clean accuracy of~\(80\%\). Note that for the validity of our certificates we assume adversaries perturb nodes after sparsification and cannot attack the sparsification itself.

\textbf{Efficient message-interception smoothing.}
Drawing Monte-Carlo samples from \(\phi\) to estimate the smoothed classifier is usually the most costly part when computing smoothing-based certificates \citep{cohen2019certified}. In \autoref{fig:archAndSparsification}~(c) we show that our certificates are much more sample efficient as we do not benefit from more than a few thousand samples from \(\phi\). This is in stark contrast to existing smoothing-based certificates for GNNs \citep{bojchevski2020efficient}. For a fair comparison, we adopt their transductive setting and compute certificates for \(\pdel=0.3\) and \(\pabl=0.85\). \citet{bojchevski2020efficient} use \(10^6\) Monte-Carlo samples for certifying test nodes on Cora-ML, which takes up to 25 minutes. In contrast, our certificates saturate already for \(2{,}000\) Monte-Carlo samples in this setting, which takes only 17 seconds (preprocessing Cora-ML takes 8 additional seconds). Our gray-box certificates are significantly more sample-efficient while also providing guarantees against much stronger adversaries. We hypotheise that our certificates saturate much faster as the certifiable radius does not depend on the inverse Gaussian CDF of the label probabilities as discussed in \autoref{sec:discussion}.

\begin{figure}[ht]
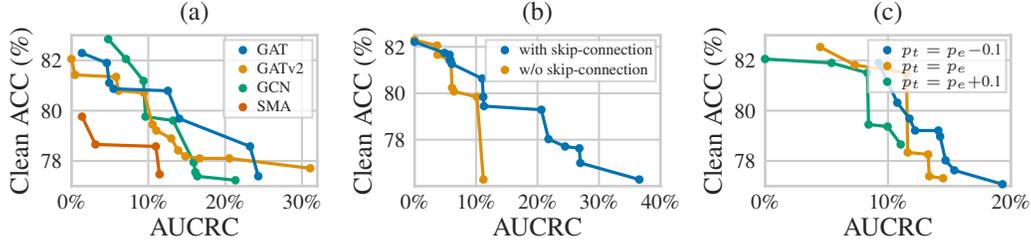

  \centering
  \begin{minipage}{0.333\textwidth}
    \centering
    \input{figures/fig34.pgf}
  \end{minipage}\hfill%
  \begin{minipage}{0.333\textwidth}
    \centering
    \input{figures/fig46.pgf}
  \end{minipage}\hfill%
  \begin{minipage}{0.333\textwidth}
    \centering
    \input{figures/fig63.pgf}
  \end{minipage}
  \caption{Second-hop attacks on Cora-ML:
  (a)~Robustness-accuracy tradeoffs for different GNN architectures. (b)~Skip-connections yield improved robustness-accuracy tradeoffs for node feature ablation smoothing. (c)~Ablating less during training yields better robustness-accuracy tradeoffs~(GAT).
  }\label{fig:naiveBaselineEfficency}%
\end{figure}

\textbf{Different classifiers.}
In \autoref{fig:naiveBaselineEfficency} (a) we compare robustness-accuracy tradeoffs for different GNNs against second-hop attacks. Attention-based message-passing GNNs \citep{velivckovic2018graph} are dominating. We hypothesize that the degree-normalization of GCN \citep{kipf2017semi} may be problematic for the performance under randomized edge deletion. Our approach may promote novel message-passing architectures, specifically designed for smoothed~classifiers.

\textbf{Skip-connections.}
With higher node feature ablation probability, more messages from the target node itself will be intercepted, which may be detrimental for the accuracy. Assuming adversaries do not attack target nodes, we can modify the architecture for improved robustness-accuracy tradeoffs (\autoref{fig:naiveBaselineEfficency}b). To this end, we forward the non-ablated input graph through the GNN \textit{without edges}, and add the resulting final representation of each node to the final representation when forwarding the (ablated) graph with graph structure. We use the same weights of the base GNN, but more complex skip-connections are straightforward. Such skip-connections yield better robustness-accuracy tradeoffs against second-hop attacks, but we also loose guarantees for the target node itself. To account for that, future work could deploy existing smoothing methods for features of target nodes~separately: e.g., if nodes represent images, we could deploy Gaussian smoothing \citep{cohen2019certified} on node features send through the skip-connection and still obtain robustness guarantees for target~nodes.

\textbf{Training-time smoothing parameters.}
In \autoref{fig:naiveBaselineEfficency} (c) we show that ablating less during training can improve the robustness-accuracy tradeoffs. Note that only inference-time smoothing parameters determine the strength of our certificates, and the probabilities \(\pdel,\pabl\) during training are just hyperparameters that we can optimize to improve the robustness-accuracy tradeoffs. In detail, we experiment with three different settings: Using the same ablation probabilities during training and inference (\(p_t=p_e\)), ablating 10\% more during training (\(p_t=p_e\)\(+\)0.1), and ablating 10\% less during training (\(p_t\)=\(p_e\)\(-\)0.1). Note that we use \(\max(\min(p_t,1),0)\) to project the training-time parameters into~\([0,1]\).

\begin{figure}[h]
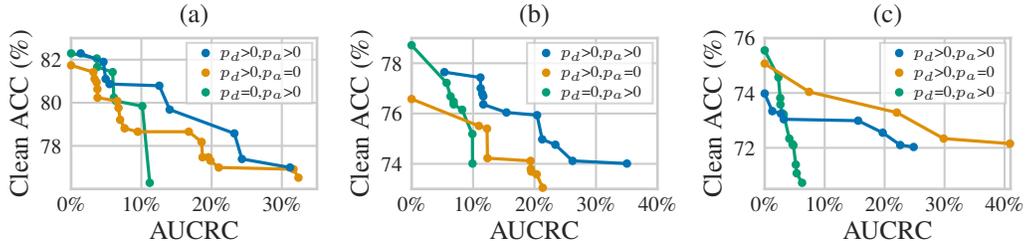

  \centering
  \begin{minipage}{0.333\textwidth}
    \centering
    \input{figures/fig22.pgf}
  \end{minipage}\hfill%
  \begin{minipage}{0.333\textwidth}
    \centering
    \input{figures/fig40.pgf}
  \end{minipage}\hfill%
  \begin{minipage}{0.333\textwidth}
    \centering
    \input{figures/fig41.pgf}
  \end{minipage}
  \caption{Robustness-accuracy tradeoffs for second-hop attacks against smoothed GAT models (without skip). Edge deletion and node ablation dominates on Cora-ML~(a) and Citeseer~(b). On PubMed~(c), edge deletion is stronger. Lines connect dominating points on the Pareto~front.
    }\label{fig:woSkip}
\end{figure}

\textbf{Robustness-accuracy.}
We compare robustness-accuracy tradeoffs of three different~settings: (1)~edge deletion and feature ablation~(\(\pdel>0\), \(\pabl>0\)), (2)~edge deletion only~(\(\pdel>0,\pabl=0\)), and (3)~feature ablation only~(\(\pdel=0,\pabl>0\)). Our experiments show that edge deletion \emph{and} feature ablation smoothing achieves significantly better robustness-accuracy tradeoffs against attribute attacks to the \emph{second-hop} neighborhood and dominates on Cora-ML and Citeseer~(\autoref{fig:woSkip}b,c). On PubMed, edge deletion smoothing dominates. More results (e.g.\ with skip-connections) in \autoref{appendix:fullresults}. 

%% file: parts/part-5-related-work.tex
\section{Related Work}

\textbf{GNN robustness.} The vast majority of GNN robustness works focus on heuristic defenses, including adversarial graph detection \citep{,zhang2020defensevgae,zhang2019comparing}; architecture modifications \citep{brody2021how,zhang2019bayesian}; robust aggregations \citep{geisler2020reliable}; robust training procedures \citep{xu2019topology,zuegner2019certifiable}, transfer learning \citep{tang2020transferring}; and graph preprocessing techniques such as edge pruning \citep{zhang2020gnnguard,wu2019adversarial}, low-rank approximations \citep{entezari2020all}, and graph anomaly detection \citep{ma2021survey}.

The effectiveness of such seemingly robust defenses on the adversarial robustness of GNNs can only be assessed against existing adversarial attacks. Heuristic defenses do not guarantee robustness, and may even be broken by stronger attacks later on \citep{mujkanovic2022_are_defenses_for_gnns_robust}. Instead, we are interested in robustness certificates that \textit{provably guarantee} the stability of predictions. However, robustness certificates for GNNs are still in their infancy \citep{GNNBook-ch8-gunnemann}:

\textbf{Certificates for GNNs.} Most certificates for GNNs are designed for specific architectures
\citep{zugner2020certifiable,jin2020certified,bojchevski2019certifiable,zuegner2019certifiable}. Despite providing provable robustness guarantees, their applicability is limited to specific architectures. \cite{bojchevski2020efficient} present the first tight and efficient smoothing-based, model-agnostic certificate for graph-structured data. However, their method comes with crucial limitations: First, their method cannot certify robustness against arbitrary feature modifications of entire nodes. Second, their black-box certificate deletes edges but completely ignores the underlying \textit{message-passing} principle. Third, their certificate requires an expensive evaluation of the smoothed classifier, which questions the practicability of their certificate beyond theoretical robustness~assessments.

Randomized ablation certificates for image classifiers \citep{levine2020robustness} are another approach for discrete data. Such certificates have already been applied to point cloud classifiers \citep{liu2021pointguard} and even for individual attribute perturbations in GNNs \citep{bojchevski2020efficient}. However, \citet{bojchevski2020efficient} show that their method outperforms such ablation certificates for individual attributes. In contrast, we propose to certify entire nodes, instead of only a few of their attributes. As already discussed, applying their ablation certificates for image classifiers directly to GNNs comes with serious shortcomings that we overcome (Section \ref{sec:certificates} and details in \autoref{appendix:comparisonLevine}).

\textbf{Gray-box certificates.} Exploiting model knowledge to derive tighter randomized smoothing certificates constitutes a widely unexplored research problem. The first works derive tighter guarantees using information about the model's gradients \citep{mohapatra2020higher,levine2020tight}. Recently proposed collective certificates \citep{schuchardt2021collective}  incorporate knowledge about the receptive fields of GNNs. Their certificates are \emph{orthogonal} to ours, and our certificates could lead to significant improvements in such collective settings, as adversaries cannot attack first-hop neighbors of all nodes~simultaneously. \citet{schuchardt2022invariance} propose tight gray-box certificates for models that are invariant to spatial transformations.

%% file: parts/part-6-conclusion.tex
\section{Conclusion}
We propose novel gray-box, message-passing aware robustness certificates for GNNs against strong threat models where adversaries can arbitrarily manipulate all features of multiple nodes in the graph. The main idea of our certificates is to intercept adversarial messages by randomly deleting edges and/or masking features of entire nodes. Our certificates are significantly stronger and more sample-efficient than existing methods. Future enhancements could smooth specific edges and nodes with different probabilities, for example to intercept messages from central nodes with higher probability. Our gray-box certificates  could lead to novel architectures, training techniques and graph preprocessing techniques to further strengthen the robustness of GNNs against adversarial examples.

%% file: parts/part-7-appendix.tex
\section*{Checklist}

\begin{enumerate}
\item For all authors...
\begin{enumerate}
  \item Do the main claims made in the abstract and introduction accurately reflect the paper's contributions and scope?
    \answerYes{All claims in abstract and introduction reflect the contributions and scope of our paper.
    We also provide a list of our core contributions directly in our introduction.}
  \item Did you describe the limitations of your work?
    \answerYes{We discuss the limitations of our approach in \autoref{sec:discussion}.}
  \item Did you discuss any potential negative societal impacts of your work?
    \answerYes{
      Without doubt, adversarial attacks can have negative impacts on the society.
      Particularly alarming are recent attacks against GNNs for more realistic threat models \citep{ma2020towards}
      and attacks that scale to large graphs \citep{geisler2021robustness}.
      Robustness certificates are tools to assess robustness
      and help to (1) better understand robustness, (2) build more robust classifiers, 
      and (3) eventually prevent adversarial attacks including their negative consequences.
      Our certificates represent a contribution towards diminishing and preventing
      potential negative impacts of adversarial attacks on the society.
      }
  \item Have you read the ethics review guidelines and ensured that your paper conforms to them?
    \answerYes{We carefully reviewed the ethics guidelines (\url{https://neurips.cc/public/EthicsGuidelines}).
    Our paper conforms to all ethic guidelines.
    Our certificates represent a contribution to prevent negative social impacts of adversarial attacks, as discussed above.}
\end{enumerate}

\item If you are including theoretical results...
\begin{enumerate}
  \item Did you state the full set of assumptions of all theoretical results?
  \answerYes{We state all assumptions of our theoretical results.}
  \item Did you include complete proofs of all theoretical results?
  \answerYes{We show all statements, including additional elaborations, in the Appendix.
  We always link theoretical results in the paper to the corresponding proofs in the Appendix.}
\end{enumerate}

\item If you ran experiments...
\begin{enumerate}
  \item Did you include the code, data, and instructions needed to reproduce the main experimental results (either in the supplemental material or as a URL)?
  \answerYes{We uploaded the code required to reproduce our main results.
  All required datasets are publicly available, 
  and can be loaded for example with PyTorch Geometric \citep{fey2019fast}.}
  \item Did you specify all the training details (e.g., data splits, hyperparameters, how they were chosen)?
    \answerYes{We describe important training details directly at the beginning of our experiment section (\autoref{sec:experiments}). 
    We further thoroughly list all training details in~\autoref{appendix:hyperparameters}.}
  \item Did you report error bars (e.g., with respect to the random seed after running experiments multiple times)?
    \answerYes{We repeat each experiment for five random splits and model initializations,
    only report averaged results, and our plots explicitly show the standard deviation over the five experiments (shaded areas in plots).
    In separate experiments, we tested our method for another set of five randomly drawn seeds,
    but we did not observe significant differences in the results.
    We also uploaded all seeds to ensure reproducibility.
    }
  \item Did you include the total amount of compute and the type of resources used (e.g., type of GPUs, internal cluster, or cloud provider)?
    \answerYes{
      We discuss the runtime for each experiment in our experiment section (\autoref{sec:experiments}).
      We conduct all experiments in an internal cluster with the following GPU type: NVIDIA GeForce GTX 1080 Ti.
    }
\end{enumerate}

\item If you are using existing assets (e.g., code, data, models) or curating/releasing new assets...
\begin{enumerate}
  \item If your work uses existing assets, did you cite the creators?
    \answerYes{Yes, we cite all authors of the assets we use.}
  \item Did you mention the license of the assets?
    \answerNA{Our datasets are well-established research datasets with MIT License or public domain.}
  \item Did you include any new assets either in the supplemental material or as a URL?
    \answerYes{We provide supplemental materials and code to reproduce our main results at \url{https://www.cs.cit.tum.de/daml/interception-smoothing}.}
  \item Did you discuss whether and how consent was obtained from people whose data you're using/curating?
    \answerNA{Our assets do not require consent.}
  \item Did you discuss whether the data you are using/curating contains personally identifiable information or offensive content?
    \answerNA{Our assets do not contain any personal, protected or offensive data.}
\end{enumerate}

\item If you used crowdsourcing or conducted research with human subjects...
\begin{enumerate}
  \item Did you include the full text of instructions given to participants and screenshots, if applicable?
    \answerNA{We do not conduct research with human subjects.}
  \item Did you describe any potential participant risks, with links to Institutional Review Board (IRB) approvals, if applicable?
    \answerNA{We do not conduct research with human subjects.}
  \item Did you include the estimated hourly wage paid to participants and the total amount spent on participant compensation?
    \answerNA{We do not conduct research with human subjects.}
\end{enumerate}

\end{enumerate}

\newpage

\section{Proofs Main Certificate (\autoref{sec:certificates})}\label{appendix:certificateProofs}

\boundingProposition{1}{}{}
\begin{proof}
  For a thorough formal proof in the context of image classifiers see \citep{levine2020robustness}.
  Here, we show the statement in the context of GNNs:
  Consider a fixed target node \(v\).
  We exploit that whenever we intercept all adversarial messages (i.e. nodes are disconnected or we mask out their features),
  the adversary cannot alter the prediction. 
  Let \(\bar{E}\) denote the event that \(v\) does not receive any message from perturbed nodes.
  Then we have for any class \(y\):
	\[
		p(f_v(\phi(G))=y \mid \bar{E}) = p(f_v(\phi(G'))=y \mid \bar{E})
	\]
  since all input representations with respect to \(G\) and \(G'\),
  which affect the prediction for~\(v\), are the same if all perturbed nodes are ablated or disconnected
  (i.e. their messages are intercepted). Multiplying with \(p(\bar{E})\) yields:
  \begin{equation}\label{eq:levineproof}
		p(f_v(\phi(G))=y \wedge \bar{E}) = p(f_v(\phi(G'))=y \wedge \bar{E})
  \end{equation}
  Following the arguments of \citep{levine2020robustness}:
  \begin{equation*}
    \begin{split}
      p_{v,y}(G) - p_{v,y}(G')
      & \stackrel{(1)}{=} p(f_v(\phi(G))=y \wedge E) + p(f_v(\phi(G))=y \wedge \bar{E}) - p_{v,y}(G') \\
      & \stackrel{(2)}{=} p(f_v(\phi(G))=y \wedge E) + p(f_v(\phi(G'))=y \wedge \bar{E}) - p_{v,y}(G') \\
      & \stackrel{(3)}{=} p(f_v(\phi(G))=y \wedge E) - p(f_v(\phi(G'))=y \wedge E) \\
      & \leq p(f_v(\phi(G))=y \wedge E) \\
      & \stackrel{(4)}{\leq} p(E) \\
    \end{split}
  \end{equation*}
  where (1) and (3) follow from the law of total probability, 
  (2) is due to inserting \autoref{eq:levineproof},
  and \((4)\) follows from \(p(A\cap B) \leq p(B)\) for any events \(A\) and \(B\).

  \noindent Analogously, \(p_{v,y}(G') - p_{v,y}(G) \leq p(E) \). Thus:
  \( |p_{v,y}(G) - p_{v,y}(G')| \leq p(E) = \Delta \)
\end{proof}

\messageInterceptionLemma{1}{}{}
\begin{proof}
  The prediction \(f_v(\phi(G))\) cannot differ from \(f_v(\phi(G'))\) if
  for all perturbed nodes \(w\in\mathbb{B}\) we have 
  (1) \(w\) is disconnected from the target node \(v\),
  or (2) the features of \(w\) are ablated.
  If the smoothing distribution \(\phi_1\) deletes an edge \((i,j)\) (that is \(\phi(\bm{A})_{ij}=0\)),
  the neighborhood \(\mathcal{N}(j)\) changes, 
  and thus messages from \(i\) to \(j\) get intercepted on all GNN~layers.
  That is, the final hidden representation \(\bm{h}^{(k)}_v\) of a target node \(v\) 
  can only be changed by some non-ablated perturbed source node \(w\)
  if there is at least one simple path from \(w\) to \(v\) of length at most \(k\)
  such that no edge on this path is deleted.
\end{proof}

\boundingTheorem{1}{}{}
\begin{proof}
  Note the difference:
  \begin{itemize}
    \item \(E\) denotes the event that at least one message from perturbed nodes reaches a target node \(v\)
    \item \( E(\bm{\rho}_v) \) denotes the event that at least one message from nodes indicated by \(\bm{\rho}_v\) reaches a target node \(v\)
  \end{itemize}
  Put differently, the maximization amounts to the additional worst-case assumption that the adversary selects
  those nodes whose messages have the highest chance of getting to the target~node.
  Importantly, we have to make this additional worst-case assumption to obtain valid robustness certificates
  for our threat model.
\end{proof}

Since the probability \(\Delta\) bounds the worst-case change \(|p_{v,y}(G) - p_{v,y}(G')|\) for all classes~\(y\),
we can utilize \(\Delta\) to construct robustness certificates:
Intuitively, \(\Delta \) bounds how much probability mass of 
the distribution \(p_{v,y}(G)\) over labels \(y\) is compromised by the worst-case adversary:
If an adversary cannot shift enough probability mass to change the majority class, our smoothed classifier is robust:
\begin{restatable}[Binary Certificate]{corollary}{theoremBinaryCert}
  Given \(\Delta\) as defined in 
  Then we can certify the robustness \(g_v(G) = g_v(G')\) for any graph \(G' \in B_{\rho}(G)\) if
  \[ p_{v,y^*}(G) - \Delta > \frac{1}{2} \]
  where \(y^* \triangleq g_v(G)\) denotes the majority class predicted by smoothed classifier \(g\).
\end{restatable}
\begin{proof}
	Recall that \(\Delta \) bounds how much probability mass of
  the distribution \(p_{v,y}(G)\) over \(y\) is compromised by the adversary.
	Let \(y^* \triangleq g(G)\) denote the majority class, that is \( p_{v,y^*}(G) > \frac{1}{2} \)
	in this binary classification setting.
	Thus, to change the majority class, the adversary needs to 
	shift enough probability mass from the majority class \(y^*\) to the other class \(1-y^*\).
	This is impossible if  \( p_{v,y^*}(G) - \Delta > \frac{1}{2} \),
	meaning the adversary cannot shift enough probability mass for a successful attack.
  Put differently, even in the worst-case that the adversary always changes the prediction whenever adversarial messages
  reach the target node, the majority class cannot be altered.
\end{proof}

\corMulticlassCert{1}{}{}
\begin{proof}
	To prove this, we utilize the same arguments as in the binary setting above.
	Here, given \(p_{v,y^*}(G) - \Delta > \max_{\tilde{y}\neq y^*} p_{v,\tilde{y}}(G) + \Delta\),
	the adversary does not control enough probability mass of \(p_{v,y}(G)\) over \(y\)
	to alter the second-best class \(\tilde{y}\) into the new majority class
  when classifying the perturbed graph \(G'\). 
\end{proof}

\corollaryPracticalCerts{2}{}{}
\begin{proof}
  We have
  \(   
    p_{v,y^*}(G) - \Delta
    \geq
    \underline{p_{v,y^*}(G)} - \overline{\Delta} 
    > \overline{p_{v,\tilde{y}}(G)} + \overline{\Delta} 
    \geq
    p_{v,\tilde{y}}(G) + \Delta
  \)
  due to the assumption
  \( 
    \underline{p_{v,y^*}(G)} - \overline{\Delta} > \overline{p_{v,\tilde{y}}(G)} + \overline{\Delta}
  \).
  The remaining claim follows from \autoref{corMulticlassCert}
  and from the fact that both bounds hold with significance level \(\alpha\).
\end{proof}

\clearpage
\section{Theoretical Connection to Randomized Ablation for Image Classifiers}\label{appendix:nodeAblation}\label{appendix:comparisonLevine}

Our gray-box certificates for GNNs are theoretically
related to the randomized ablation black-box certificates for image classifiers.
In this section we thoroughly analyze the differences with more technical insights
and carefully discuss how our certificates go beyond theirs.
Specifically, we show that our gray-box certificates yield stronger guarantees,
and are provably tighter even in the special case without additional edge deletion smoothing.
In the following we introduce their certificate again, discuss the differences to our certificate, 
and eventually prove that our guarantees are tighter.

\paragraph{Randomized Ablation.}
\citet{levine2020robustness} introduce randomized ablation for image classifiers as follows:
They define the space 
\(B(n,k) \triangleq \{ M : M \in \mathcal{P}(\{1, \ldots, n\}) \wedge |M| = k \}\)
of all pixel-subsets with exactly \(k\) of \(n\) total pixels
(\(\mathcal{P}\) denoting the power set here).
Then, their smoothing distribution ablates all but \(k\) pixels 
in a uniformly drawn subset \(M\in B(n,k)\).
They define \(\Delta^L\) as the probability to \emph{keep} (not ablate)
perturbed pixels in the image under this smoothing distribution.
Assuming \(\rho\) perturbed pixels in an image:
\[
  \Delta^L = 1-\frac{\binom{n-\rho}{k}}{\binom{n}{k}}
\]
\paragraph{Discussion.}
There are various ways of applying such black-box certificates for image classifiers 
to certify the robustness of GNNs.
One way is to use them to certify threat models
where adversaries control individual attributes all over the graph \citep{bojchevski2020efficient}.
We are interested in certifying robustness to adversaries
that control all features of entire nodes in the graph instead.
However, applying the smoothing distribution of \citet{levine2020robustness}
for certifying robustness to our threat model (that is by ablating entire node vectors)
comes with several deficiencies,
as their smoothing distribution is specifically designed for image classifiers.
Most importantly, applying their certificate for image classifiers to GNNs
results in black-box certificates that completely ignore the underlying message-passing principle.

In contrast, we propose gray-box certificates --
we partially open the black-box 
and consider the underlying \emph{message-passing} principle
and paths in the graph, that is \(\bm{A}\) and~\(\bm{A}^2\).
This comes with two crucial advantages
as we show experimentally in \autoref{sec:experiments}:
First, additionally deleting edges leads to significantly better robustness guarantees
for attacks against more distant nodes.
Second, our certificates become increasingly stronger
for sparser graphs
(while their certificate applied to GNNs remains unchanged 
as it ignores graph structure).

\subsection{Special Case of Node Feature Ablation Smoothing}
Notably, our certificates are provably tighter even without edge deletion smoothing.
Specifically, we formally show the difference 
between our \(\Delta\) for node feature ablation smoothing 
and \(\Delta^L\) of \citet{levine2020robustness} when naively applying their approach to GNNs
by randomly ablating features of entire nodes (instead of pixels in an image).
Specifically, while their smoothing distribution
samples exactly \(k\) out of \(n\) nodes not to ablate (to keep),
our smoothing distribution samples \(k\) out of \(n\) nodes \textit{in expectation}.
This eventually leads to \(\Delta < \Delta^L\).
We start by characterizing our certificate for node ablation smoothing:

\propNodeAblationFirst{2}{}{}
\begin{proof}
  Recall the definition of the probability \(\Delta\):
  \(E\) denotes the event that at least one perturbed message reaches a target node \(v\),
  and \(\Delta \triangleq p(E)\).
  When only ablating nodes (\(\pdel=0\)),
  all nodes are equally important for the prediction \(f_v(\phi(G))\),
  since messages are only intercepted in the input layer,
  not during the message-passing itself.

  We therefore do not have an optimization problem as in \autoref{boundingTheorem}.
  Instead, the probability \(\Delta\) to receive perturbed messages 
  is just the probability that at least one perturbed node is not ablated.
  Further, the \textit{complementary event} denotes that all \(\rho\) perturbed nodes are ablated,
  whose probability is just \(\pabl^\rho\).
  Thus \(\Delta = 1-\pabl^\rho\).
\end{proof}

\clearpage
Moreover, the multiplicative bound is tight
in the special case of node ablation smoothing:
\begin{restatable}{proposition}{propNodeAblationThird}
  For \(\pdel=0\), the multiplicative bound is tight \(\overline{\Delta}_M = \Delta\).
\end{restatable}
\begin{proof}
  We have
  \[
    \overline{\Delta}_i \stackrel{(1)}{=} \left[ 1-\prod_{q\in{P_{wv}^k}} \left( 1- (1-\pdel)^{|q|} \right) \right](1-\pabl) 
    \stackrel{(2)}{=} 1-\pabl
  \]
  where \((1)\) is by definition, and \((2)\) due to our assumption \(\pdel=0\).
  Therefore:
  \[ 
    \overline{\Delta}_M 
    = 1-\prod_{i=1}^{\rho} (1-\overline{\Delta}_i)
    = 1-\prod_{i=1}^{\rho} \pabl
    = 1-\pabl^\rho
    = \Delta
   \] 
   where the first equality is due to definition again, and the last equality follows from \autoref{propNodeAblationFirst}.
\end{proof}

\begin{proposition}[Tighter guarantees]\label{prop:cmplevine}
  Given adversarial budget \(\rho > 1\).
  Further assume \(k>0\).
  Let \(\Delta^L\) denote the bounding constant for the smoothing distribution proposed by \citet{levine2020robustness}.
  Then \(\Delta < \Delta^L\).
\end{proposition}
\begin{proof}
  Recall that due to uniform ablation we have (compare \citet{levine2020robustness}):
  \[
    \Delta^L = 1-\frac{\binom{n-\rho}{k}}{\binom{n}{k}}
  \]
  To compare this to our \(\Delta = 1-\pabl^\rho\) of \autoref{propNodeAblationFirst},
  we first need to introduce \(k\) and~\(n\).
  We note that \(\pabl\) is the probability to ablate a single node.
  We thus have \(\pabl = 1-\frac{k}{n}\),
  where \(\frac{k}{n}\) amounts to the probability to ``keep'' (not ablate) a node.
  In this setting, we keep \(n\frac{k}{n} = k\) nodes in expectation.
  We therefore have:
  \[
    \Delta = 1-\pabl^\rho = 1-\left(1-\frac{k}{n}\right)^\rho 
  \]
  We observe:
  \[ 
    \frac{\binom{n-\rho}{k}}{\binom{n}{k}}
    = \frac{(n-\rho)!(n-k)!}{n!(n-\rho-k)!}
    = \prod_{i=0}^{\rho-1}\frac{n-k-i}{n-i}
    \stackrel{(1)}{<} \left(\frac{n-k}{n}\right)^\rho
    = \left(1-\frac{k}{n}\right)^\rho
  \]
  where \((1)\) is due to the mediant inequality (\(\rho > 1\) and \(k>0\)): 
  \[ \forall y<x \; \forall i>0:\quad \frac{y-i}{x-i} < \frac{y}{x} \]
  We conclude that \( \Delta < \Delta^L\).
\end{proof}

\noindent The difference decreases for larger n, 
but our smoothing distribution is significantly better 
for small graphs/receptive fields:
For example, for \(n=10\) and \(k=1\) (i.e. \(\pabl=0.9\)), the largest certifiable radius 
with our method is \(6\), but only \(4\) using their certificate.

In detail, there are two ways of applying their method 
for image classifiers to certify robustness of GNNs against adversaries 
that control all features of entire nodes in the graph:
by ablating all features of \(k\) out of \(n\) uniformly chosen nodes 
(1)~\emph{in the entire graph}, or (2)~\emph{locally in each receptive~field}.

\paragraph{Global randomized ablation.}
Assume we uniformly ablate all features of \(k\) out of \(n\) nodes \emph{in the entire graph}.
If the number of nodes \(n\) in the graph is large,
the difference between \(\Delta\) and \(\Delta^L\) is small.
Still, the resulting black-box certificates only hold globally, 
not locally in the receptive fields. 
Such certificates ignore the receptive fields, 
specifically that most nodes in the graph may not even be connected to the target node.
For example, in the most extreme case of \(\bm{A} = \bm{0}\) 
(meaning receptive fields only consist of target nodes),
their certificate applied to GNNs remains entirely unchanged due to the black-box nature.
In contrast,  
our gray-box certificates guarantee robustness 
for any \(\rho\) (excluding target nodes) in this case 
(cf. normalized robustness in~\autoref{sec:experiments}).

\begin{figure}[h]
  \centering
  \begin{minipage}{0.5\textwidth}
    \centering
    \input{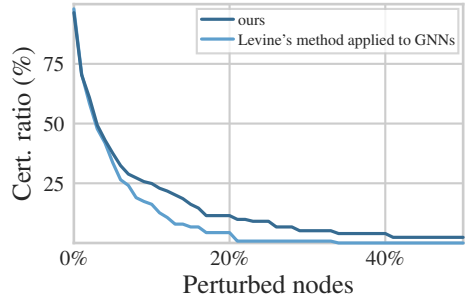}
  \end{minipage}\hfill%
  \caption{
    Given \(\pabl=0.72\), we compare our certificate against the
    certificate proposed by \cite{levine2020robustness}
    by applying their smoothing distribution for image classifiers
    to GNNs (distance \(\geq\) 1, with skip-connection).
    We locally choose \(k = \lfloor(n-1)*\pabl\rfloor\) nodes not to ablate
    -- where \(n-1\) is the number of nodes in each receptive field,
    excluding the target node.
    Our certificates are experimentally stronger even 
    without additional edge deletion.
  }\label{fig:appendix:levineforGNNs}
\end{figure}

\paragraph{Local randomized ablation.}
To remedy the black-box nature of their approach, one can obtain local guarantees by ablating all features of \(k\) out of 
the \(n\) nodes \emph{locally in the receptive field} of a target node.
However, our message-interception certificates 
are significantly tighter even without edge deletion smoothing as
receptive fields are typically small.
We demonstrate this in \autoref{fig:appendix:levineforGNNs}
where our approach yields significantly stronger guarantees in practice
(since \autoref{propNodeAblationFirst} makes a significant difference).

Note that when applying their approach to GNNs by ablating nodes locally,
one also needs to consider each receptive field individually
and cannot use full-batch training/inference as usually implemented for GNNs.
Our message-interception certificates are easier to implement and more efficient 
as we obtain local guarantees without considering and processing 
all receptive fields separately.

\clearpage
\section{Closed-form via Inclusion-exclusion Principle}\label{appendix:ieprinciple}
Recall that \( E(\bm{\rho}_v) \) describes the event that \(v\) receives messages from 
\textit{any} attacked node indicated by the adversarial budget vector \(\bm{\rho}_v\in\{0,1\}^n\).
Computing the probability \(p\left(E({\bm{\rho}_v})\right)\) using edge deletion probability \(\pdel\)
and node feature ablation probability \(\pabl\) is challenging as it involves evaluating 
the inclusion-exclusion formula. We formalize this expensive closed-form solution in the following:
Let \(E_w\) denote the probability to receive a message from node \(w\),
and let \(\mathcal{P}\) indicate all simple paths from any perturbed \(w\) with \(\bm{\rho}_v(w)=1\) to target node \(v\).
Further, let \(Y_{i}\) denote the probability to receive a message via path \(i\in\mathcal{P}\).
Then we have:
\[
  p\left(E({\bm{\rho}_v})\right)
   = p\left(\bigvee_{\bm{\rho}_v(w)=1} E_w\right)
   = p\left(\bigvee_{i\in\mathcal{P}} Y_{i}\right)
\]
since the probability to receive a message from any attacked node 
equals the probability to receive a message from any path \(i\) from an attacked node to the target node.
We now apply the inclusion-exclusion~principle:
\begin{equation}\label{eq:ieprinciple}
   p\left(\bigvee_{i\in\mathcal{P}} Y_{i}\right)
   = \sum_{k=1}^{|\mathcal{P}|} \left( (-1)^{k-1} 
        \sum_{\substack{\mathcal{I}\subseteq \mathcal{P} \\ |\mathcal{I}|=k}} p\left(\bigwedge_{i\in \mathcal{I}} Y_i\right)\right)
\end{equation}
The remaining probability can be expressed as follows:
The probability to receive messages via \textit{all} paths indicated by \(\mathcal{I}\)
is the probability that (1) all edges on those paths are not deleted,
and (2) the corresponding source nodes of the paths are not ablated.
Therefore:
\begin{equation}\label{eq:ieprinciple2}
  p\left(\bigwedge_{i\in\mathcal{I}} Y_i\right) = (1-\pdel)^a (1-\pabl)^b
\end{equation}
where \(a\) denotes the number of (unique) edges on all paths indicated by \(\mathcal{I}\),
and \(b\) the number of (unique) source nodes of the paths indicated by \(\mathcal{I}\).
Note that the above derivation assumes 
that the target node \(v\) is not controlled by the adversary.
In such a case (\(\bm{\rho}_v(v)=1\)),
we have \(p(E_v) = 1-\pabl\) (since we always receive messages from non-ablated target nodes)
and:
\[
  p\left(E({\bm{\rho}_v})\right)
   = p\left(\bigvee_{i\in\mathcal{P}} Y_{i} 
            \bigvee E_v
            \right)
  \quad\quad
  p\left(\bigwedge_{i\in\mathcal{I}} Y_i \bigwedge E_v\right) 
  \stackrel{(1)}{=}p\left(\bigwedge_{i\in\mathcal{I}} Y_i \right) p(E_v)
  \quad\quad 
\]
where (1) is due to independence.

There are different ways that take additional information into account
to derive faster ways of computing \(p\left(E({\bm{\rho}_v})\right)\),
for example by exploiting that the receptive fields are trees 
with the target node \(v\) as root (compare \autoref{appendix:trees}).
In general, however, computing \autoref{eq:ieprinciple} is expensive since 
we have to evaluate \autoref{eq:ieprinciple2} exactly \(2^{|\mathcal{P}|}\) times.

\clearpage
\section{Tree-shaped Receptive Fields}\label{appendix:trees}

Given fixed \(\bm{\rho}_v\in\{0,1\}^n\) that indicates nodes controlled by the adversary.
Recall that \( E(\bm{\rho}_v) \) describes the event that \(v\) receives 
at least one messages from 
\textit{any} attacked node indicated by the adversarial budget vector \(\bm{\rho}_v\in\{0,1\}^n\).
If the receptive field for target node \(v\) is a tree, we can compute \(\Delta\) of \autoref{boundingTheorem} exactly.
Specifically, we first provide a recursive formula to compute \(p\left(E({\bm{\rho}_v})\right)\)
and then show that the worst-case selection of nodes by the adversary is straightforward.

We introduce the following random variables to better describe the recursion: 
\begin{itemize}
  \item Let \(R_i\) denote the event that root node \(i\) receives an adversarial message.
  \item Let \(A_i\) denote the event that the features of node \(i\) are ablated.
  \item Let \(D_i\) denote the event that root \(i\) receives an adversarial message 
  via any of its adjacent subtrees \(j\in\mathcal{B}\)  (``branches'').
  \item Let \(B_j\) further denote the event that we receive an adversarial message via branch \(j\).
\end{itemize}
The main idea is that branches in a tree are independent:
\begin{theorem}
We start the recursion with the target node \(v\) to compute \(p(R_v)\) 
while following edges away from the node \((j, v)\) (against their direction). 
Then the following recursive equation computes \(p\left(E({\bm{\rho}_v})\right)\)
for tree-shaped receptive fields:
\begin{equation*}
  p(R_i) \triangleq 
  \begin{cases}
    1-\pabl (1-p(D_i)) & \textrm{if } \bm{\rho}_v(i) = 1 \\
    p(D_i) & \textrm{else}
  \end{cases}
\end{equation*}
with
\[ p(D_i) \triangleq 1-\prod_{(j,i)} \left( 1-p(B_j) \right) \quad\quad p(B_j) \triangleq (1-\pdel)p(R_j) \]
\end{theorem}
\begin{proof}
  We show the three equations consecutively:
  \begin{enumerate}
    \item For \(p(R_i)\): If root \(i\) is not controlled by the adversary,
    then the probability to receive an adversarial message is just the probability that we receive such a
    message via any of its adjacent subtrees, that is \(p(R_i) = p(D_i)\).
    If root \(i\) is controlled by the adversary (\(\bm{\rho}_v(i) = 1\)),
    we can exploit independence between 
    edge deletion smoothing \(\phi_1\) and node feature ablation smoothing \(\phi_2\):
    \[ 
      p(R_i) 
      = p(\bar{A}_i \vee D_i ) 
      = 1-p(A_i \wedge \bar{D}_i) 
      \stackrel{(1)}{=} 1-p(A_i)p(\bar{D}_i) 
      = 1-p(A_i)(1-p(D_i))
    \]
    where \((1)\) is due to independence.
    Since the probability that we do not receive any adversarial message from root \(i\) is 
    the probability that the features of root \(i\) are ablated: \(p(A_i)=\pabl\).
    We therefore have: \( p(R_i) = 1-\pabl(1-p(D_i)) \).
    \item For \(p(D_i)\): For the probability
    that root \(i\) receives an adversarial message 
    via any of its adjacent branches \(j\in\mathcal{B}\), we exploit independence between branches 
    (which we can do since we have trees):
    \[
      p(D_i) 
      = p\left(\bigvee_{j\in\mathcal{B}} B_j\right) 
      = 1-p\left(\bigwedge_{j\in\mathcal{B}} \bar{B}_j\right)
      \stackrel{(1)}{=} 1-\prod_{j\in\mathcal{B}} p(\bar{B}_j)
      = 1-\prod_{j\in\mathcal{B}} \left( 1-p(B_j) \right)
    \]
    where \((1)\) is due to independence.
    \item For \(p(B_j)\): The probability to receive a message via branch \(j\) is the
    probability that the edge from branch \(j\) to root \(i\) is not deleted (\(1-\pdel\))
    times the probability that we receive a message via the next root \(j\) (recursive call).
  \end{enumerate}
  For leaves we have \(\mathcal{B}=\emptyset\) and thus the product over \(j\in\mathcal{B}\) is \(1\),
  that is \(p(D_i) = 0\) for all~leaves.
\end{proof}

Interestingly, we can reconstruct the following special cases:

\textbf{Special case of edge deletion smoothing.} Assume \(\pabl=0\). Then we directly see that \(p(R_i)=1\) if root \(i\) is controlled by the adversary. This means that the adversary controls the entire sub-tree if the root node is already attacked. Put differently, the adversary does not need to control more parts of the tree to change the prediction if the adversary already controls the root.

\textbf{Special case of node feature ablation smoothing.} Assume \(\pdel=0\). Then we can directly see that resolving the recursion just multiplies the node feature ablation probabilities \(\pabl\) and we get \( p\left(E({\bm{\rho}_v})\right) = 1-\pabl^{\rho}\) for \(\rho=||\bm{\rho}_v||_0\). This matches the special case already discussed in \autoref{propNodeAblationFirst}.

\textbf{Worst-case selection of nodes.}
Recall that our certificates are conservative and assume the additional worst-case that
the adversary attacks those nodes in the receptive field 
that maximize the probability that the target node receives 
a message from attacked nodes (maximization in \autoref{boundingTheorem}).
This additional assumption is required to obtain valid certificates.
Notably, this worst-case adversary is straightforward for trees:
First, an adversary would always prefer closer nodes over more distant nodes
to maximize the probability that messages are getting through.
Second, an adversary would always distribute its budget over different branches
to exploit independence between branches, which also maximizes
the probability that messages are getting through (also compare \autoref{appendix:boundProofs}).

\begin{figure}
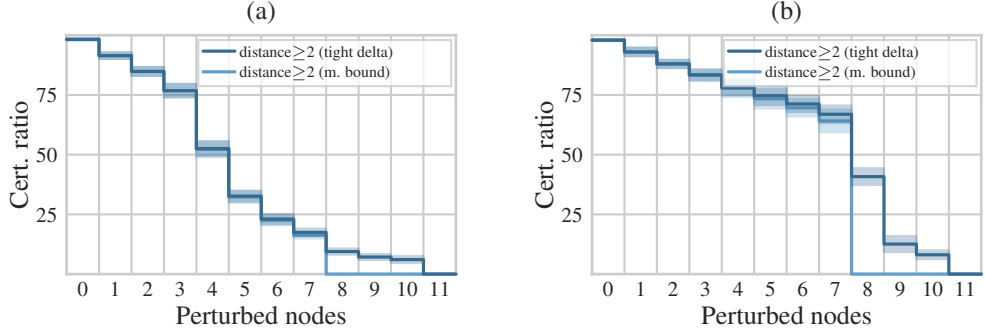

  \centering
  \begin{minipage}{0.5\textwidth}
    \centering
    \input{figures/fig38.pgf}
  \end{minipage}\hfill%
  \begin{minipage}{0.5\textwidth}
    \centering
    \input{figures/fig39.pgf}
  \end{minipage}
  \caption{Comparing multiplicative bound and tight tree bound (distance at least 2).
  (a) Tree-certificate only for tree-shaped receptive fields.
  (b) Sparsifying all receptive fields into trees.
  }\label{fig:trees2}
\end{figure}

\textbf{Experiments.}
We find that computing \(\Delta\) tight for tree-shaped receptive fields can 
increase the certifiable radius in practice (compare \autoref{fig:trees2}).
Interestingly, \(25\%\) of nodes in Cora-ML have receptive fields that are trees (considering 2-layer GNNs).
We apply our recursive scheme above to compute tight certificates in two settings:
First, we only compute tight certificates for the nodes whose receptive fields are trees.
Second, we apply sparsification that successively deletes edges in the graph 
until the receptive fields of all test nodes are trees.
In detail, we train GAT models on Cora-ML and apply sparsification at test time.
We use the skip-connection, train with \(\pabl=0.68\), \(\pdel=0.02\) 
and compute certificates with \(\pabl=0.79\), \(\pdel=0.36\).
Without sparsification we achieve clean accuracies of \(79\%\) on average,
and \(77\%\) when applying sparsification at test time.

In practice, we found that the gain in computing \(\Delta\) exactly may be rather small,
as adversaries typically distribute their budget to different branches
to increase the probability that their messages arrive.
This means adversaries maximize independencies between edges.
In other words, the multiplicative bound is already quite strong in practice,
and specifically tight until the degree of the node (given that each first-hop neighbor has at least one child).

\clearpage
\section{Proofs of \autoref{sec:practical}}\label{appendix:boundProofs}

\begin{figure}[!ht]
  \centering
  \begin{minipage}{0.5\textwidth}
      \centering
      \includegraphics[width=0.5\textwidth]{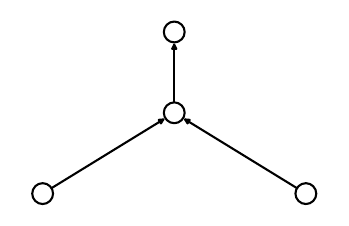}
  \end{minipage}\hfill%
  \begin{minipage}{0.5\textwidth}
    \centering
      \includegraphics[width=0.5\textwidth]{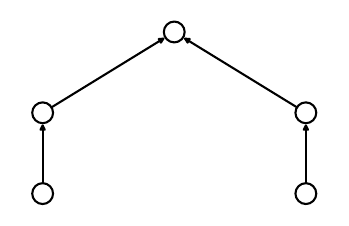}
  \end{minipage}
  \caption{Visualization of two dependent (left) and independent paths (right).
  When randomly deleting edges with the same edge deletion probability \(\pdel\),
  the probability that all messages from both source nodes 
  are intercepted is lower when the paths are independent (more possibilities for the message to get through).
  }\label{fig:pathIndependence}
\end{figure}

We first prove a more general claim that we can use to prove the multiplicative bounds of 
\autoref{singleSourceMB} and \autoref{theoremMultiplicativeBound}.
Let \(X_i\) denote the event that target node \(v\) receives a message
via any path \(s\) in a set of paths \(S_i\) such that 
all paths start at an arbitrary source node and end at target node \(v\).
Intuitively, it is more likely to receive at least one messages via \(S_i\) 
\textit{and} one message via \(S_j\)
when there are shared edges, compared to when we assume their paths were independent.
Put differently, the probability that all messages from all paths are intercepted 
is higher when paths are dependent (cf. \autoref{fig:pathIndependence}).
More formally:

\begin{theorem}\label{theorem:pathIndependence}
  For two arbitrary sets \(S_i\) and \(S_j\) of simple paths with the same target node \(v\) we have
  \[
    p(\overline{X}_i)p(\overline{X}_j) \leq p(\overline{X}_i \wedge \overline{X}_j)
  \]
  under the smoothing distribution \(\phi_1\) for edge deletion.
\end{theorem}
\textit{Proof.}
  We are interested in the probability that all messages via all paths are intercepted.
  Consider the following two possibilities:
  \begin{enumerate}
    \item The paths in \(S_i\) and the paths in \(S_j\) are (pairwise) independent,
    meaning there are no edges that appear on both - on a path \(s_i\in S_i\) and on a path \(s_j\in S_j\).

    In this case we have 
    \( p(\overline{X}_i \wedge \overline{X}_j) = p(\overline{X}_i)p(\overline{X}_j) \)
    due to independence.

    \item Consider the scenario where there are at least two dependent paths that share a common edge.
    If we assume they were independent, there would be more possibilities how a message can get through
    than there actually are. In other words, assuming independence results in lower probability 
    that all messages via both sets get intercepted. 

    Thus \( p(\overline{X}_i)p(\overline{X}_j) < p(\overline{X}_i \wedge \overline{X}_j) \).\hfill\(\square\)
  \end{enumerate}%
Consider the following definition of positively associated random variables \citep{esary1967association}.
\begin{definition}
  We call a random vector \(\bm{x} = (X_1, \ldots, X_n)\) positively associated if
  \[Cov(\phi(\bm{x}), \psi(\bm{x}))) \geq 0\]
  for all non-decreasing, element-wise functions \(\phi\), \(\psi\) such that 
  second moments of \(\psi(\bm{x})\) and \(\phi(\bm{y})\)~exist.
\end{definition}
The concept of positively associated random variables is for example used in physical statistics \citep{goldstein2018stein}.
We can use this concept here to prove multiplicative bounds:
\begin{corollary}
  The random vector \( \bm{x} = (X_1, \ldots, X_n) \) is positively associated.
\end{corollary}
\begin{proof}
  Due to \autoref{theorem:pathIndependence} we have \(p(\overline{X}_i)p(\overline{X}_j) \leq p(\overline{X}_i \wedge \overline{X}_j)\) and thus
  \begin{equation*}
    \begin{split}
      \Rightarrow & 
      \mathbb{E}[\overline{X}_i]\mathbb{E}[\overline{X}_j] \leq \mathbb{E}[\overline{X}_i\overline{X}_j] \\
      \Rightarrow &
      \mathbb{E}[\overline{X}_i\overline{X}_j] - \mathbb{E}[\overline{X}_i]E[\overline{X}_j] \geq 0 \\
      \Rightarrow & 
      Cov(\overline{X}_i, \overline{X}_j) \geq 0 \\
    \end{split}
  \end{equation*}%
  since \(\overline{X}_i\) and \(\overline{X}_j\) are binary random variables.

  Thus, the elements of the covariance matrix are non-negative: \(Cov(\bar{\bm{x}}, \bar{\bm{x}}) \geq 0\) (variance is always non-negative). 
  According to Theorem 4.2 in \cite{esary1967association},
  \(\bar{\bm{x}}\) is positively associated.
  Since \(\bar{\bm{x}}\) is positively associated, it follows from (BP1) 
  in \cite{esary1967association} that
  \(\bm{x}\) is positively associated.
\end{proof}

\begin{restatable}{proposition}{multiplicativeProp}
  Given random variables \(X_i\) as defined above. Then:
  \[
  1-p\left(\bigwedge_{i=1}^{n} \overline{X}_i\right) 
    \leq 1-\prod_{i=1}^{n} p\left(\overline{X}_i\right) 
  \]
\end{restatable}
\begin{proof}
  Since \(\bm{x}\) and \(\bar{\bm{x}}\) are positively associated random variables, we can use
  Theorem 4.1 in \citep{esary1967association} and conclude that 
  \[
    p\left(\bigwedge_{i=1}^{n} \overline{X}_i\right) \geq \prod_{i=1}^{n} p\left(\overline{X}_i\right)
    \Leftrightarrow
    1-p\left(\bigwedge_{i=1}^{n} \overline{X}_i\right) 
    \leq 1-\prod_{i=1}^{n} p\left(\overline{X}_i\right) 
  \]
\end{proof}

\singleSourceMB{2}{}{}
\begin{proof}
  Note in the special case of the target node \(v=w\) we just have \(\Delta_w = 1-\pabl\),
  since the features \(\bm{x}_v\) of the target node \(v\) are used for the prediction independent of any~edges.

  For any \(w\neq v\) in the receptive field:
  Let \(E_w\) denote the event that the target node \(v\) receives messages from node \(w\),
  and \( \Delta_w \triangleq p(E_w)\).
  We further introduce \(A_w\) for the event that the features of node \(w\) are ablated,
  and \(D_w\) for the event that \(v\) receives at least one messages from~\(w\). Then we have:
  \[ \Delta_w = p(E_w) = p(\bar{A}_w \wedge D_w) \stackrel{(1)}{=} p(\bar{A}_w) p(D_w) = (1-\pabl) p(D_w)\]
  where \((1)\) holds since the two smoothing distributions for node feature ablation and edge deletion are independent. 
  We continue with \(p(D_w)\).
  Therefore, recall that \(\mathcal{P}\triangleq\mathcal{P}_{wv}^k\) denotes the set of simple paths from \(w\) to~\(v\).
  Further, let \(p(q)\) for simple path \(q\in\mathcal{P}\) denote 
  the probability that \(v\) receives a message via path \(q\).
  Clearly, a message ``arrives'' only via path \(q\) if none of the edges on that path is deleted,
  that is when the node is connected via path \(q\).
  Since the deletion of edges is independent, \(p(q) = (1-\pdel)^{|q|} \),
  where \(|q|\) denotes the number of edges on the simple path \(q\).
  We derive:
  \[ 
    p(D_i) 
    = p\left( \bigvee_{q\in\mathcal{P}} q \right) 
    = 1-p\left( \bigwedge_{q\in\mathcal{P}} \overline{q} \right)
  \]
  We can use positive association to conclude
  \[
    1-p\left( \bigwedge_{q\in\mathcal{P}} \overline{q} \right)
    \stackrel{(1)}{\leq} 1-\prod_{q\in\mathcal{P}} p\left(\overline{q} \right)
  \]
  where \((1)\) follows from \autoref{thmt@@multiplicativeProp}.
  Finally, we resolve the remaining terms:
  \[
    1-\prod_{q\in\mathcal{P}} p\left(\overline{q} \right)
    = 1-\prod_{q\in\mathcal{P}} \left( 1- p\left(q \right) \right)
    = 1-\prod_{q\in\mathcal{P}} \left( 1-(1-\pdel)^{|q|} \right)
  \]
  Due to \((1)\) above, we finally get \( \Delta_w \leq \overline{\Delta}_w \),
  where the inequality becomes an equality if all paths are independent (that is the paths do not share edges).
\end{proof}

\begin{restatable}{proposition}{propTightSingleSourceDelta}
  We have \(\Delta_w = \overline{\Delta}_w\) for \(\ell\)-layer GNNs with \(\ell \leq 2\).
\end{restatable}
\begin{proof}
  For \(\ell\)-layer GNNs with \(\ell \leq 2\), all paths from a single source to the target node are independent.
\end{proof}

\theoremMultiplicativeBound{3}{}{}
\begin{proof}
  We recall from \autoref{boundingTheorem}:
  \[
    \Delta = \max_{||\bm{\rho}_v||_1\leq \rho} p\left(E({\bm{\rho}_v)}\right) 
  \]
  where \( E(\bm{\rho}_v) \) describes the event that 
  target node \(v\) receives \textit{messages} from any attacked node indicated by \(\bm{\rho}_v\).
  Recall that \(E_w\) denotes the event that the prediction for target node \(v\) is based on information of node \(w\)
  in the receptive field. We further have \( \Delta_w \triangleq p(E_w)\). Then:
  \[
    p\left(E({\bm{\rho}_v})\right) 
    = p\left(\bigvee_{\bm{\rho}_v(w)=1} E_w\right) 
    = 1-p\left(\bigwedge_{\bm{\rho}_v(w)=1} \bar{E}_w\right) 
  \]
  where we can apply \autoref{thmt@@multiplicativeProp}
  and use the assumption that paths from several source nodes to the target were independent
  to obtain an upper bound:
  \[
    1-p\left(\bigwedge_{\bm{\rho}_v(w)=1} \bar{E}_w\right) 
    \leq 1-\prod_{\bm{\rho}_v(w)=1} p\left(\bar{E}_w\right) 
  \]
  Further resolving the terms yields:
  \[
    1-\prod_{\bm{\rho}_v(w)=1} p\left(\bar{E}_w\right) 
    = 1-\prod_{\bm{\rho}_v(w)=1} \left( 1-p\left(E_w\right) \right)
    = 1-\prod_{\bm{\rho}_v(w)=1} \left( 1-\Delta_w \right)
  \]
  Since the above equations hold for any fixed \(\bm{\rho}_v\):
  \[
    \Delta 
    = \max_{||\bm{\rho}_v||_1\leq \rho} p\left(E({\bm{\rho}_v)}\right) 
    \leq \max_{||\bm{\rho}_v||_1\leq \rho} 1-\prod_{\bm{\rho}_v(w)=1} \left( 1-\Delta_w \right)
  \]
  Assume we have ordered \(\Delta_w\) so that 
  \(\Delta_{i} \geq \Delta_{i+1}\) for all \(i\in\{1, \ldots, \rho\}\). Then:
  \[
    \max_{||\bm{\rho}_v||_1\leq \rho} 1-\prod_{\bm{\rho}_v(w)=1} \left( 1-\Delta_w \right)
    = 1-\prod_{i=1}^{\rho} \left( 1-\Delta_i \right)
    = \overline{\Delta}_M
  \]
\end{proof}
Note that instead of \(\Delta_w\) we can alternatively use upper bounds \(\overline{\Delta}_w\),
which yields an even looser upper bound on~\(\Delta\) since
\[
    1-\prod_{i=1}^{\rho} \left( 1-\Delta_i \right)
    \leq 
    1-\prod_{i=1}^{\rho} \left( 1-\overline{\Delta}_i \right)
\]

\clearpage
\section{Approximation Error}\label{appendix:approximationError}

Notably, the multiplicative bound derived above is tighter than the following union bound:
\begin{proposition}[Union Bound]
  Given monotonously decreasing \(\Delta_i\) such that \(\Delta_{i} \geq \Delta_{i+1}\).
  Then we have \(\Delta \leq \overline{\Delta}_U\) for
  \[
    \overline{\Delta}_U \triangleq \sum_{i=1}^{\rho} \Delta_i
  \]
\end{proposition}
\begin{proof}
  \[
    p\left(E({\bm{\rho}_v)}\right) 
    = p\left(\bigvee_{\bm{\rho}_v(w)=1} E_w\right) 
    \leq \sum_{\bm{\rho}_v(w)=1} p\left(E_w\right)
    = \sum_{\bm{\rho}_v(w)=1} \Delta_w
  \]\[
    \Delta 
    = \max_{||\bm{\rho}_v||_1\leq \rho} p\left(E({\bm{\rho}_v})\right) 
    \leq \max_{||\bm{\rho}_v||_1\leq \rho} \sum_{\bm{\rho}_v(w)=1} p\left(E_w\right)
    = \sum_{i=1}^{\rho} \Delta_i
  \]
\end{proof}

\begin{figure}[h]
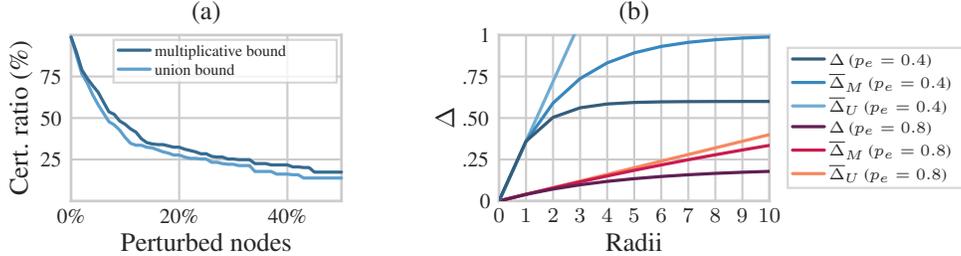

  \centering
  \begin{minipage}{0.4\textwidth}
    \centering
    \input{figures/fig23.pgf}
  \end{minipage}\hfill%
  \begin{minipage}{0.6\textwidth}
    \centering
    \input{figures/fig19.pgf}
  \end{minipage}\hfill%
  \caption{
   (a) Multiplicative bound is tighter than union bound
   and provides stronger guarantees
   (Smoothed GAT model on Cora-ML with \(\pabl=0.85, \pdel=0\)).
   (b) Constructed example: All path share the same bottleneck edge:
   Comparing the tight \(\Delta\) against the union bound \(\overline{\Delta}_U\) 
   and the multiplicative bound \(\overline{\Delta}_M\)
   for different edge deletion probabilities \(\pdel\).
   The multiplicative bound is tighter than the union bound, which can grow larger than \(1\).
  }\label{fig:appendix:multiplicativeVsUnion}
\end{figure}

The union bound is quite loose, not a probability and can even grow larger than \(1\). 
We show the difference in practice \autoref{fig:appendix:multiplicativeVsUnion} (a).
We also discuss the approximation error between the upper bounds
\(\overline{\Delta}_U\), \(\overline{\Delta}_M\)
and the tight \(\Delta\) for the following constructed example
where all paths are dependent:
We assume a setting where an adversary attacks only second-hop neighbors
that are connected to the target node via the same direct neighbor of the target node.
With \(\pabl=0\) we have 
\(
  \Delta = (1-\pdel)(1-\pdel^\rho)
\)
since we only receive a message if the bottleneck edge is not ablated,
and at least one edge of the attacked second-hop nodes is not ablated
(which is the complementary probability of all second-hop edges are ablated).
In this constructed case, all paths are dependent as they share the bottleneck edge.
We show how the upper bounds compare to the tight \(\Delta\) for different
edge deletion probabilities~\(\pdel\) in \autoref{fig:appendix:multiplicativeVsUnion} (b). Note that the example is constructed and worst-case adversaries aim at maximizing independencies by choosing nodes without bottleneck edges (in which case the multiplicative bound is a strong bound in practice).

\clearpage
\section{Hyperparameters}\label{appendix:hyperparameters}

We implement certificates for directed and undirected graphs.
For our main experiments (\autoref{sec:experiments}), however,
we follow the standard procedure and prepocess 
all graphs into undirected graphs, only consider the largest connected component,
and binarize node features.
We compute simple paths using a modified depth first search.
All datasets are included in PyTorch Geometric
\citep{fey2019fast}.%
\footnote{\url{https://pytorch-geometric.readthedocs.io}}
We train models full-batch using Adam
(learning rate = 0.001,
 \(\beta_1  = 0.9\),
 \(\beta_2  = 0.999\),
 \(\epsilon = 10^{-08}\),
 weight decay = \(5*10^{-04}\))
for \(1{,}000\) epochs with early stopping after 50 epochs.
We use a dropout of \(0.8\) on the feature matrix \(\bm{X}\)
and on the attention coefficients.
During training, we sample a different graph from \(\phi(G)\) each epoch.
Each sampled graph contains nodes with features replaced by the ablation representation \(\bm{t}\).
We implement \(\bm{t}\) as a parameter of our models:
We initialize \(\bm{t}\) using Xavier initialization and we optimize 
\(\bm{t}\) as we optimize the GNN weights during training.
We implement all models for two message-passing layers.
We use 8 heads and 8 hidden channels
for GAT and GATv2 \citep{velivckovic2018graph,brody2021how};
64 hidden channels for GCN \citep{kipf2017semi};
and we use \(k=64\) and temperature=\(1.0\) for SMA \citep{geisler2021robustness}. We use the ReLU activation function for the skip-connection. For GDC sparsification, we set the sparsification threshold of GDC to \(\epsilon=0.022\), and ignore edge attributes resulting from GDC preprocessing.

\textbf{Training-time smoothing parameters.}
We also delete edges and ablate node features during training 
(using different probabilities \(\pdel\) and \(\pabl\) during training and inference). Specifically, we train models presented in \autoref{sec:experiments} as follows:
In \autoref{fig:skipConnection} (a,b) we show results for 
\(\pdel=0.01, \pabl=0.6\) during training (and 
\(\pdel=0.31, \pabl=0.794\) during inference and certification).
In \autoref{fig:archAndSparsification} (a,b) we use
\(\pdel=0, \pabl=0.59\) during training (and 
\(\pdel=0.31, \pabl=0.71\) during inference and certification).
In \autoref{fig:archAndSparsification} (c) we use the same probabilities \(\pdel, \pabl\) during training and inference.

In our experiments (\autoref{sec:experiments}), we also randomly sample different probabilities for training and inference
to explore the joint parameter space of the training-time and inference-time smoothing parameters.
That is, our search space is \([0,1]^4\)
when sampling different probabilities from \([0,1]\)
for the Pareto-plots in \autoref{fig:woSkip}
and \autoref{appendix:fullresults} (we sample separately for training and inference).

\section{Detailed Results}\label{appendix:fullresults}\label{appendix:arxiv}\label{appendix:alpha}

\begin{figure}[!ht]
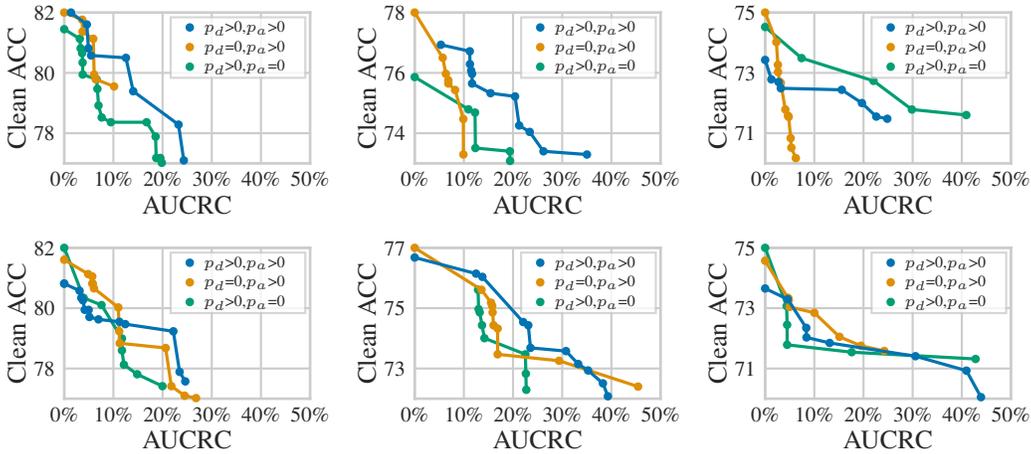

  \centering
  \begin{minipage}{0.333\textwidth}
    \centering
    \input{figures/full_results/cora_ml_GAT_False_2.pgf}
  \end{minipage}\hfill%
  \begin{minipage}{0.333\textwidth}
    \centering
    \input{figures/full_results/citeseer_GAT_False_2.pgf}
  \end{minipage}\hfill%
  \begin{minipage}{0.333\textwidth}
    \centering
    \input{figures/full_results/pubmed_GAT_False_2.pgf}
  \end{minipage}
  \begin{minipage}{0.333\textwidth}
    \centering
    \input{figures/full_results/cora_ml_GAT_True_2.pgf}
  \end{minipage}\hfill%
  \begin{minipage}{0.333\textwidth}
    \centering
    \input{figures/full_results/citeseer_GAT_True_2.pgf}
  \end{minipage}\hfill%
  \begin{minipage}{0.333\textwidth}
    \centering
    \input{figures/full_results/pubmed_GAT_True_2.pgf}
  \end{minipage}
  \caption{
    Robustness-accuracy tradeoffs for second-hop attacks
    against \emph{smoothed GAT} on Cora-ML, Citeseer and PubMed (columns).
    Top row without skip-connection, bottom row with skip-connection.
    Lines connect dominating points on the Pareto front.
  }\label{fig:appendix:paretoGAT}
\end{figure}

\begin{figure}[ht]
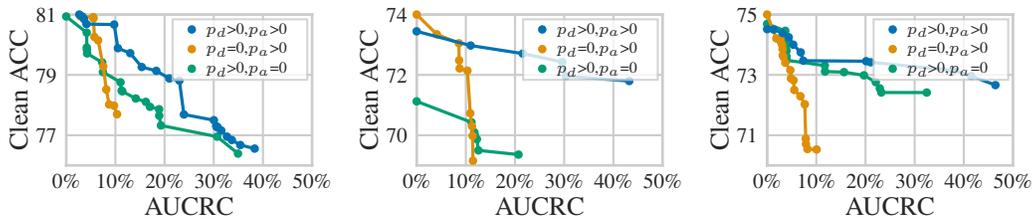

  \centering
  \begin{minipage}{0.333\textwidth}
    \centering
    \input{figures/full_results_transductive/cora_ml_GAT_False_2.pgf}
  \end{minipage}\hfill%
  \begin{minipage}{0.333\textwidth}
    \centering
    \input{figures/full_results_transductive/citeseer_GAT_False_2.pgf}
  \end{minipage}\hfill%
  \begin{minipage}{0.333\textwidth}
    \centering
    \input{figures/full_results_transductive/pubmed_GAT_False_2.pgf}
  \end{minipage}
  \caption{
    Transductive learning setting:
    Robustness-accuracy tradeoffs for second-hop attacks
    against \emph{smoothed GAT} on Cora-ML, Citeseer and PubMed.
    Experiments without skip-connection.
  }\label{fig:transductiveSetting}
\end{figure}

\begin{figure}[!ht]
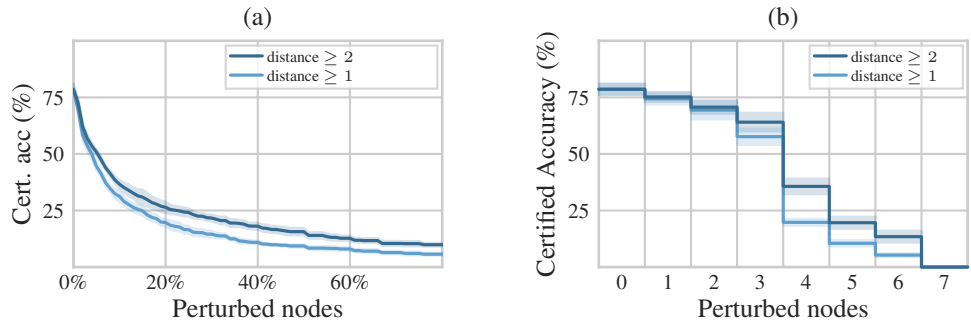

  \centering
  \begin{minipage}{0.5\textwidth}
      \centering
      \input{figures/fig30.pgf}
  \end{minipage}\hfill%
  \begin{minipage}{0.5\textwidth}
      \centering
      \input{figures/fig51.pgf}
  \end{minipage}
  \caption{Certified accuracies for the setting of \autoref{fig:skipConnection} -- Smoothed GAT on Cora-ML:
  (a)~Robustness at different distances (\(\pdel\)=\(0.31\), \(\pabl\)=\(0.794\), with skip-connection, ACC=\(0.79\)).}\label{fig:appendix:certACCs}
\end{figure}

We report certified accuracies in \autoref{fig:appendix:certACCs}
for the corresponding certified ratios in \autoref{fig:skipConnection}.
Moreover, we provide detailed results for the datasets Cora-ML, Citeseer, and PubMed.
We show results for second-hop attacks against (1)
smoothed GAT models in \autoref{fig:appendix:paretoGAT},
(2) smoothed GATv2 models in \autoref{fig:appendix:paretoGATv2},
(3) smoothed GCN models in \autoref{fig:appendix:paretoGCN}, and
(4) smoothed SMA models in \autoref{fig:appendix:paretoSMA}.
We run 1{,}000 experiments for each combination,
drawing random deletion and ablation probabilities from \([0,1]\)
for each experiment (sampling separately for training and inference).
Lines connect dominating points on the Pareto front.
Comparing results with and without skip-connection we observe that 
skip-connections allow higher node feature ablation probabilities
while~retaining high accuracy,
which can yield better robustness-accuracy tradeoffs.
Moreover, as discussed in \autoref{sec:experiments}, evaluating certificates in transductive settings comes with serious shortcomings. We nevertheless report such results in \autoref{fig:transductiveSetting} for a smoothed GAT model.

\begin{figure}[!ht]
  \centering
  \input{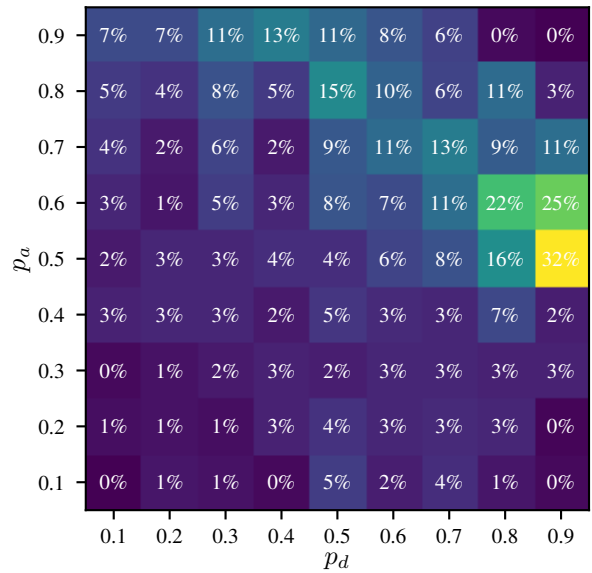}
  \caption{
    Abstained ratios of smoothed GAT 
    models trained on Cora-ML
    for different edge deletion probabilities \(\pdel\)
    and node feature ablation probabilities \(\pabl\).
  }\label{fig:appendix:ablation}
\end{figure}

\textbf{Abstained predictions.}
Our smoothed classifier abstains from predicting if 
\(\underline{p_{v,y^*}(G)} \leq \overline{p_{v,\tilde{y}}(G)}\).
We show the ratio of abstained predictions for smoothed GAT 
models trained on Cora-ML in \autoref{fig:appendix:ablation}
for different edge deletion probabilities \(\pdel\)
and node feature ablation probabilities \(\pabl\).
We use the same ablation probability during training and inference
for this specific experiment.
We observe that our smoothed classifier abstains
for rather large probabilities.
Future work could introduce novel architectures and 
training techniques to further diminish 
the effect of abstained predictions.

\begin{figure}[!ht]
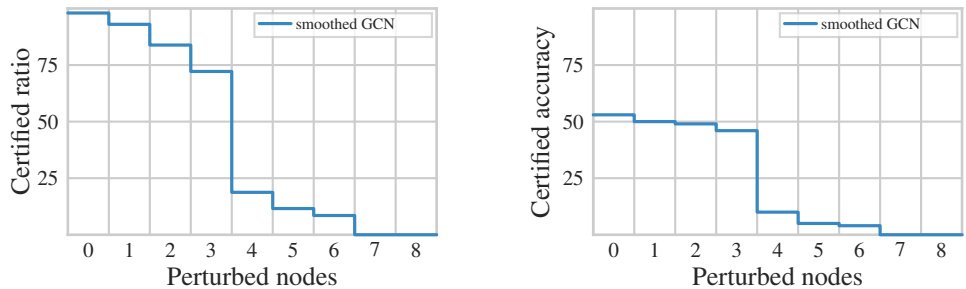

  \centering
  \begin{minipage}{0.5\textwidth}
      \centering
      \input{figures/fig47.pgf}
  \end{minipage}\hfill%
  \begin{minipage}{0.5\textwidth}
      \centering
      \input{figures/fig49.pgf}
  \end{minipage}
  \caption{Certified ratio and accuracy for smoothed two-layer GCN on ogbn-arxiv.
  We certify 100 randomly selected test nodes in the graph.
  Certificates for nodes with distance 2 to the target node.
  }\label{fig:appendix:arxivResults}
\end{figure}

\textbf{Experiments on ogbn-arxiv.}
We run additional experiments and compute certificates for
the larger graph ogbn-arixv with 169{,}343 nodes,
128 attributes and  40 classes \citep{hu2020ogb}.
We adopt their transductive setting,
implement two-layer smoothed GCNs with skip-connection
and compute certificates for 100 randomly chosen test nodes.
In \autoref{fig:appendix:arxivResults} we show results for 
\(\pdel=0.1, \pabl=0.4\) during training, and 
\(\pdel=0.3, \pabl\)=\(0.8\) during inference and certification.
Notably, we can certify GNNs for such large graphs.
However, our approach only achieves \(53\%\) clean accuracy in this setting.
Future work could develop novel architectures and training procedures 
to improve clean accuracy under our smoothing~distribution.

\textbf{Experiments with different confidence levels.}
We conduct additional experiments with varying confidence levels \(\alpha\) 
and Monte-Carlo samples. 
We observe strong guarantees for even smaller confidence levels,
requiring little computational efforts.
The underlying reason for this is that the theoretical largest certifiable radius 
of our certificates is bounded, only determined by the edge deletion probability~\(\pdel\) 
and node feature ablation probability~\(\pabl\), and therefore cannot increase by changing~\(\alpha\).
Our certificates are thus less sensitive to changes in \(\alpha\) 
compared to Neyman-Pearson-based certificates \citep{bojchevski2020efficient}.

\begin{figure}[ht]
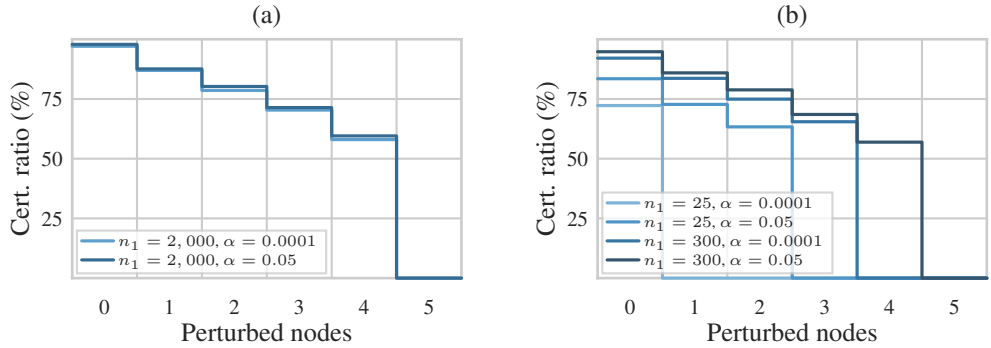

  \centering
  \begin{minipage}{0.5\textwidth}
    \centering
    \input{figures/fig52.pgf}
  \end{minipage}\hfill%
  \begin{minipage}{0.5\textwidth}
    \centering
    \input{figures/fig53.pgf}
  \end{minipage}
  \caption{
    Certified ratio of smoothed GAT on Cora-ML
    (\(\pabl=0.84\), \(\pdel=0\), with skip-connection)
    for different confidence levels \(\alpha\) 
    and number of Monte-Carlo samples \(n_1\).
    The difference in robustness is already considerably small 
    for just 2,000~samples.
  }\label{fig:appendix:alphaExperiments}
\end{figure}

\begin{figure}[!ht]
  \centering
  \input{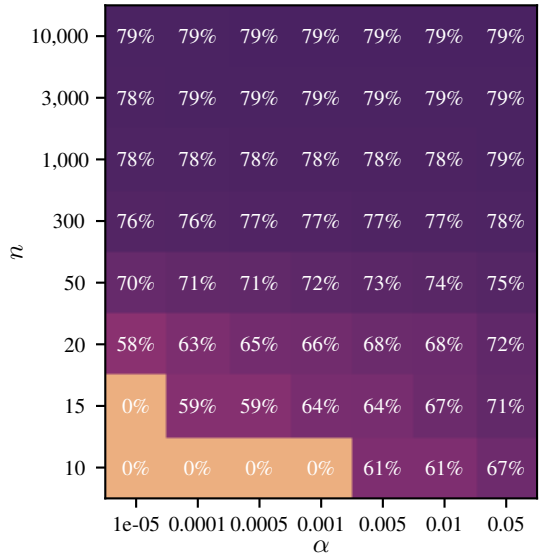}
  \caption{
    Clean accuracy of smoothed GAT on Cora-ML
    (\(\pabl=0.84\), \(\pdel=0\), with skip-connection).
    for varying number of confidence levels \(\alpha\) and 
    Monte-Carlo samples \(n\).
    For \(\alpha=0.05\) the clean accuracy is high 
    for just \(1,000\) samples.
    For smaller \(\alpha\), the certification accuracy decreases only slightly.
    Drawing more than \(3,000\) samples is not necessary except for 
    extremely small confidence levels such as \(\alpha=0.00001\).
  }\label{fig:appendix:clean_acc_heatmap}
\end{figure}

In fact, the difference in certifiable robustness 
for \(\alpha=0.05\) and \(\alpha=0.0001\) is already extremely small 
when drawing just \(2,000\) Monte-Carlo samples (\autoref{fig:appendix:alphaExperiments} a).
We only observe differences in robustness for considerably 
small amounts of Monte-Carlo samples (\autoref{fig:appendix:alphaExperiments} b).
Drawing 2,000 samples takes only 12 seconds on Cora-ML on average.
This is significantly faster compared to all previous probabilistic certificates 
for GNNs that use up to \(10^6\) Monte-Carlo samples (compare \citep{bojchevski2020efficient}).
In additional experiments, we also found that the classification accuracy is high 
for just a few thousand Monte-Carlo samples (\autoref{fig:appendix:clean_acc_heatmap}).

\begin{figure}[!ht]
  \centering
  \begin{minipage}{0.333\textwidth}
    \centering
    \input{figures/full_results/cora_ml_GATv2_False_2.pgf}
  \end{minipage}\hfill%
  \begin{minipage}{0.333\textwidth}
    \centering
    \input{figures/full_results/citeseer_GATv2_False_2.pgf}
  \end{minipage}\hfill%
  \begin{minipage}{0.333\textwidth}
    \centering
    \input{figures/full_results/pubmed_GATv2_False_2.pgf}
  \end{minipage}
  \begin{minipage}{0.333\textwidth}
    \centering
    \input{figures/full_results/cora_ml_GATv2_True_2.pgf}
  \end{minipage}\hfill%
  \begin{minipage}{0.333\textwidth}
    \centering
    \input{figures/full_results/citeseer_GATv2_True_2.pgf}
  \end{minipage}\hfill%
  \begin{minipage}{0.333\textwidth}
    \centering
    \input{figures/full_results/pubmed_GATv2_True_2.pgf}
  \end{minipage}
  \caption{
    Robustness-accuracy tradeoffs for second-hop attacks
    against \emph{smoothed GATv2} on Cora-ML, Citeseer and PubMed (columns).
    Top row without skip, bottom row with skip-connection.
  }\label{fig:appendix:paretoGATv2}
\end{figure}

\begin{figure}[!ht]
  \centering
  \begin{minipage}{0.333\textwidth}
    \centering
    \input{figures/full_results/cora_ml_GCN_False_2.pgf}
  \end{minipage}\hfill%
  \begin{minipage}{0.333\textwidth}
    \centering
    \input{figures/full_results/citeseer_GCN_False_2.pgf}
  \end{minipage}\hfill%
  \begin{minipage}{0.333\textwidth}
    \centering
    \input{figures/full_results/pubmed_GCN_False_2.pgf}
  \end{minipage}
  \begin{minipage}{0.333\textwidth}
    \centering
    \input{figures/full_results/cora_ml_GCN_True_2.pgf}
  \end{minipage}\hfill%
  \begin{minipage}{0.333\textwidth}
    \centering
    \input{figures/full_results/citeseer_GCN_True_2.pgf}
  \end{minipage}\hfill%
  \begin{minipage}{0.333\textwidth}
    \centering
    \input{figures/full_results/pubmed_GCN_True_2.pgf}
  \end{minipage}
  \caption{
    Robustness-accuracy tradeoffs for second-hop attacks
    against \emph{smoothed GCN} on Cora-ML, Citeseer and PubMed (columns).
    Top row without skip-connection, bottom row with skip-connection.
  }\label{fig:appendix:paretoGCN}
\end{figure}

\begin{figure}[!ht]
  \centering
  \begin{minipage}{0.333\textwidth}
    \centering
    \input{figures/full_results/cora_ml_SMA_False_2.pgf}
  \end{minipage}\hfill%
  \begin{minipage}{0.333\textwidth}
    \centering
    \input{figures/full_results/citeseer_SMA_False_2.pgf}
  \end{minipage}\hfill%
  \begin{minipage}{0.333\textwidth}
    \centering
    \input{figures/full_results/pubmed_SMA_False_2.pgf}
  \end{minipage}
  \begin{minipage}{0.333\textwidth}
    \centering
    \input{figures/full_results/cora_ml_SMA_True_2.pgf}
  \end{minipage}\hfill%
  \begin{minipage}{0.333\textwidth}
    \centering
    \input{figures/full_results/citeseer_SMA_True_2.pgf}
  \end{minipage}\hfill%
  \begin{minipage}{0.333\textwidth}
    \centering
    \input{figures/full_results/pubmed_SMA_True_2.pgf}
  \end{minipage}
  \caption{
    Robustness-accuracy tradeoffs for second-hop attacks
    against \emph{smoothed SMA} on Cora-ML, Citeseer and PubMed (columns).
    Top row without skip-connection, bottom row with skip-connection.
  }\label{fig:appendix:paretoSMA}
\end{figure}

\clearpage

\clearpage
\section{On Neyman-Pearson and Ablation Certificates}\label{appendix:npandablation}

There are currently two types of randomized smoothing certificates for discrete data:
The certificates of \citet{lee2019tight} and \citet{bojchevski2020efficient}
are based on the Neyman-Pearson Lemma \citep{neyman1933ix},
and we therefore call them Neyman-Pearson-based certificates.
The other certificates are ablation-based \citep{levine2020robustness,levine2020Derandomized,liu2021pointguard}.
We show that largest certifiable radius of ablation-based certificates is bounded indepdentent of the classifier, which is not the case for Neyman-Pearson-based certificates (see discussion in \autoref{sec:discussion}).

In ablation-based certificates, the bounding constant \(\Delta\) determines the probability mass of 
the distribution \(p_{v,y}(G)\) over labels \(y\) that the worst-case adversary controls.
This probability mass \(\Delta\) is independent of the classifier \(f\) and distribution \(p_{v,y}(G)\) 
and solely determined by the smoothing distribution.
Although the final certificates still depend on the classifier \(f\),
the largest certifiable radius of such ablation-based certificates is bounded 
as we show for our interception smoothing~certificates:

Note again that \(\Delta\) does not depend on the base GNN \(f\):
the probability to receive at least one message from a perturbed node is only characterized 
by the number of perturbed nodes~\(\rho\), and the probabilities \(\pdel\) for edge deletion and \(\pabl\) for node ablation.
Moreover, \(\Delta\) is monotonously increasing in~\(\rho\),
since the probability to receive messages from perturb nodes increases the more nodes adversaries~control.
Interestingly, since \(\Delta\) is monotonously increasing in \(\rho\),
there exists a largest certifiable radius that
depends on the graph structure and changes for each target node (assuming fixed \(p_d,p_a\)).
In the special case of node ablation smoothing,
we can directly determine the largest certifiable~radius:%
\propNodeAblationSecond{3}{}{}
\begin{proof}
Due to \autoref{thmt@@theoremBinaryCert} and \autoref{corMulticlassCert},
we only get certificates if \(\Delta < \frac{1}{2}\),
i.e. the adversary should not control more than half of the distribution \(p_{v,y}(G)\) over \(y\).
Thus:
\[ 
  \Delta < \frac{1}{2} 
  \stackrel{(1)}{\Leftrightarrow}
  1-\pabl^\rho < \frac{1}{2} 
  \Leftrightarrow 
  \pabl^\rho > \frac{1}{2} 
  \Leftrightarrow 
  \pabl > \sqrt[\rho]{0.5}
\]
since the root is monotonously increasing and \(\pabl>0\).
Further, \((1)\) stems from \autoref{propNodeAblationFirst}.
Thus we need an ablation probability of at least larger than \(\sqrt[\rho]{0.5}\) to certify a radius of~\(\rho\).
\end{proof}

\begin{figure}
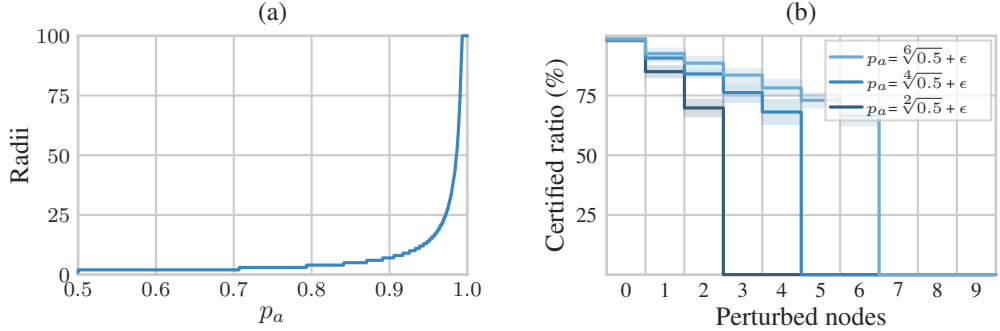

  \centering
  \begin{minipage}{0.5\textwidth}
    \centering
    \input{figures/fig18.pgf}
  \end{minipage}\hfill%
  \begin{minipage}{0.5\textwidth}
    \centering
    \input{figures/fig8.pgf}
  \end{minipage}
  \caption{
    Visualizing \autoref{propNodeAblationSecond}.
    (a) Theoretically maximally certifiable radius for given node ablation probability \(\pabl\).
    (b) Certified ratio of smoothed GAT trained on CoraML for different
    node ablation probabilities (\(\pdel=0\), \(\epsilon=0.01\)). 
    Note: \(\sqrt[2]{0.5}\approx0.71\),  \(\sqrt[4]{0.5}\approx0.84\)~and~\(\sqrt[6]{0.5}\approx0.89\).
  }\label{fig:exponentialPlot}
\end{figure}
\noindent \autoref{propNodeAblationSecond} allows us to directly determine 
the largest certifiable radius for given~\(\pabl\).
We visualize this largest radius for different ablation probabilities in \autoref{fig:exponentialPlot}~(a).
Theoretically, we can only certify large radii for relatively large ablation probabilities:
For example, to theoretically certify a radius of \(10\), we already need an ablation probability of more than 
\(\sqrt[10]{0.5}\approx 0.933\). 
\autoref{propNodeAblationSecond} implies that we cannot certify any radius for ablation probabilities \(\pabl \leq 0.5\)
(cf. \autoref{fig:SingleSourceDelta}).
Moreover, we can certify a radius of only \(1\) for ablation probabilities between
\(\sqrt[1]{0.5}=0.5\) and \(\sqrt[2]{0.5}\approx0.707\).
Note, however, that this is only a theoretical consideration and 
that the certificate also depends on the label probabilities
\(p_{v,y^*}(G)\) and \(p_{v,\tilde{y}}(G)\) in practice (\autoref{fig:exponentialPlot} b),
where we observe that the certified ratio drops to zero when the largest certifiable radius is~passed.

\clearpage

\section{Message-passing-aware Derandomization}\label{appendix:derandomization}

As discussed in \autoref{sec:discussion}, 
our certificates are probabilistic and hold with a certain confidence level \(\alpha\). Here we present alternative, deterministic certificates using a simplified smoothing distribution that just deletes nodes instead of ablating their features. We believe that future work can build upon it towards even more efficient and scalable derandomization schemes. Specifically, our derandomized certificates come with the following advantages:  First, they are deterministic, exact certificates and hold independent of a confidence level. Second, the smoothed classifier never abstains from making a prediction (we resolve draws by whatever index comes first).
Third, with more computation time we obtain more derandomized certificates. This is in continuation to probabilistic certificates that can be improved using more Monte-Carlo samples \citep{cohen2019certified}.

\textbf{Simplified smoothing distribution.}
We define a smoothed classifier that classifies node \(v\) in \(G\) as follows:
Consider a retention constant \(k\in\mathbb{N}\)
that represents the number of nodes not deleted (retained) in the receptive field. Then the smoothed classifier \(g\) predicts class~\(y\) with the largest probability \(p_{v,y}(G)\) that \(f\) classifies \(v\) as \(y\) under uniform deletion of all but \(k\) nodes:
\begin{equation*}
    g_v(G) \triangleq \arg\max_y p_{v,y}(G)
    \quad\quad
    p_{v,y}(G) \triangleq p_{\mathcal{K}\sim \mathcal{U}(d,k)}(f(\mathcal{R}_{\mathcal{K}}) = y)
\end{equation*}
where \(\mathcal{R}_\mathcal{K}\) encodes the deletion of all nodes in the receptive field of target node \(v\) except those indexed by \(\mathcal{K}\),
and \(f(\mathcal{R}_\mathcal{K})\) denotes the predicted class of \(f\) 
for target node \(v\) given ablated graph \(\mathcal{R}_\mathcal{K}\) 
(omitting \(v\) for conciseness).
We further denote the indexing of nodes \(\mathcal{K}\) as follows:
Define the set of all \(k\) unique indices in \([d] \triangleq \{1, \ldots, d\}\) including \(0\) as
\(B(d,k) = \{ \{0\}\cup M : M \in \mathcal{P}([d]) \wedge |M| = k \}\),
where \(\mathcal{P}\) denotes the power set (w.l.o.g. we index target nodes as 0).
For example, \(\mathcal{K}=\{0,1,3,6\} \in B(d,k)\) for retention constant \(k=3\) and receptive field size \(d=10\).
Note that \(|\mathcal{K}|=k+1\) for \(\mathcal{K}\in B(d,k)\) but \(|B(d,k)| = \binom{d}{k}\)
since we never delete the target node.
Finally, let \(\mathcal{U}(d,k)\) denote the uniform distribution over \(B(d,k)\).

\begin{figure}[h]
  \centering
  \begin{minipage}{0.24\textwidth}
      \centering
      (1)
      \includegraphics[width=1\textwidth]{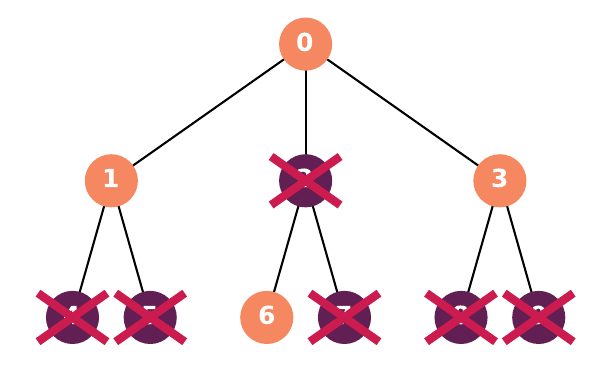}
  \end{minipage}\hfill%
  \begin{minipage}{0.24\textwidth}
      \centering
      (2)
      \includegraphics[width=1\textwidth]{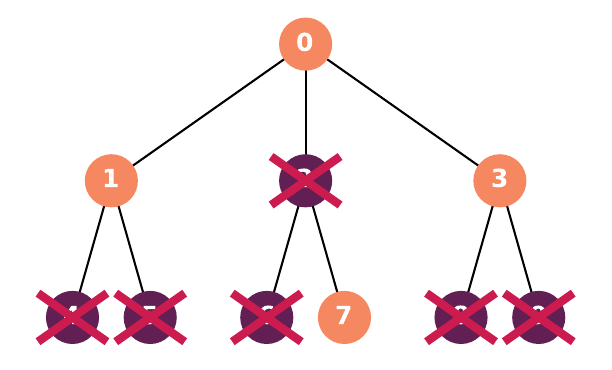}
  \end{minipage}\hfill%
  \begin{minipage}{0.24\textwidth}
      \centering
      (3)
      \includegraphics[width=1\textwidth]{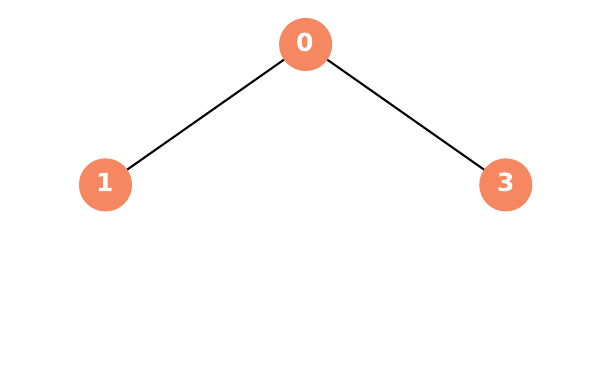}
  \end{minipage}
  \begin{minipage}{0.24\textwidth}
      \centering
      (4)
      \includegraphics[width=1\textwidth]{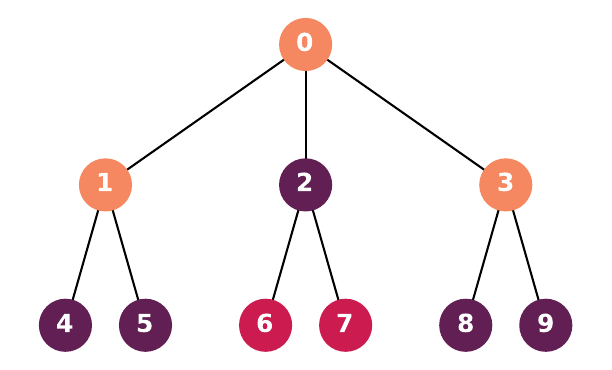}
  \end{minipage}
  \caption{
      Given a receptive field with 10 nodes, target node 0 and \(k=3\). 
      (1) If we keep nodes \(\mathcal{K}=\{0,1,3,6\}\) and delete all other nodes, node 6 is disconnected.
      (2) If we keep nodes \(\mathcal{K}=\{0,1,3,7\}\) and delete all other nodes, node 7 is disconnected.
      (3) In both cases, only the nodes \(\mathcal{S}(\mathcal{K})=\{0,1,3\}\) affect the prediction.
      (4) In the algorithm: Given \(\mathcal{S}=\{0,1,3\}\) with neighborhood \(\mathcal{N}_\mathcal{S} = \{2,4,5,8,9\}\).
      Choosing \(k+1-|\mathcal{S}|=1\) further nodes,
      we find that \(\mathcal{S}\) is a reduced representative \(\mathcal{S}(\mathcal{K})\) since there are 
      \(|V_v| - |\mathcal{N}_\mathcal{S}| - |\mathcal{S}| = 10 - 5 - 3 = 2\) nodes~to~choose~from~(6~and~7).
  }
  \label{fig:eqc}
\end{figure}

Computing \(p_{v,y^*}(G)\) and \(p_{v,\tilde{y}}(G)\) exactly is challenging. One naive approach would be to simply iterate over the support of the smoothing distribution (all possible node deletions). For small receptive fields, the number of possible combinations to sample \(k\) out of \(d\) nodes may be small, allowing us to enumerate all possibilities.
However, this may be infeasible for larger receptive fields. Still, similar to how we use the message-passing structure for certification, we can also leverage it here to partition the support of the simplified smoothing distribution into a smaller number of equivalence classes.

Specifically, we observe: First, when uniformly deleting nodes in the receptive field, some of the remaining nodes \(\mathcal{K}\) may be disconnected from the target node. Moreover, disconnected nodes will not affect the prediction for the target node. Second, several possibilities for \(\mathcal{K}\)
may share the same nodes that are still connected to \(v\) (see examples in \autoref{fig:eqc}). This means that different possibilities for \(\mathcal{K}\) will lead to the same prediction by \(f\), but the full enumeration of all possibilities is suboptimal: We wish to avoid redundant evaluations since the evaluation of the base classifier \(f\) may be costly. 

We observe that the connectivity explained above induces an equivalence relation:
All sampled nodes \(\mathcal{K}\) that share the same nodes connected to \(v\)
can be grouped into equivalence classes~\([\mathcal{K}]\).
For any representative \(\mathcal{K}\) of \([\mathcal{K}]\) we denote
the nodes still connected to \(v\) as \(\mathcal{S}(\mathcal{K})\).
We call \(\mathcal{S}(\mathcal{K})\) a reduced representative, 
since it represents a reduced form of \(\mathcal{K}\) and only
contains the nodes from which the target node will receive messages.
Note that \(\mathcal{S}(\mathcal{K})\) is unique for all representatives \(\mathcal{K}\).

Formally, given receptive field \(\mathcal{R}\) with \(d+1\) nodes
and index \(\mathcal{K}\in B(d,k)\) of \(k+1\) nodes.
Consider the subgraph \(\mathcal{R}_\mathcal{K}\) induced by \(\mathcal{K}\).
We observe that not necessarily all nodes in \(\mathcal{R}_\mathcal{K}\) have to be connected to the target node.
Thus, different \(\mathcal{K}\in B(d,k)\) will result in same prediction of the base classifier.
Let \(\mathcal{S}(\mathcal{K}) \subseteq \mathcal{K}\)
denote all nodes indexed by \(\mathcal{K}\) without the disconnected nodes.
Put differently, \(\mathcal{S}(\mathcal{K})\) stands for nodes still connected to the target node
(see example in \autoref{fig:eqc}). Then:

\begin{restatable}{proposition}{propositionEqclasses}\label{prop:Eqclasses}
    The definition of \(\mathcal{S}(\mathcal{K})\) induces an equivalence relation \(\sim\) over
    \(B(d,k)\) given by
    \(
        \mathcal{K} \sim \mathcal{K'} 
        \Leftrightarrow
        \mathcal{S}(\mathcal{K})
        =
        \mathcal{S}(\mathcal{K'})
     \)
    and eq. classes \([\mathcal{K}] := \{\mathcal{K}' \in B(d,k) : \mathcal{K} \sim \mathcal{K}' \}\)
    for \(\mathcal{K}\in B(d,k)\).
\end{restatable}
\begin{proof}
    Reflexivity, symmetry and transitivity hold by the definition of sets.
\end{proof}

The equivalence relation \(\sim\) partitions \(B(d,k)\) into disjoint equivalence classes, denoted by the quotient set
\(B(d,k)/\sim\; \triangleq \{ [\mathcal{K}] \mid \mathcal{K} \in B(d,k) \}\).
The set \(\mathcal{S}(\mathcal{K})\) is uniquely defined
for each equivalence class \([\mathcal{K}]\) in \(B(d,k)/\sim\).
We therefore call \(\mathcal{S}(\mathcal{K})\) with 
\(1 \leq |\mathcal{S}(\mathcal{K})| \leq k+1\) the \textbf{reduced representative}
of \([\mathcal{K}]\).
Note that we have \(|\mathcal{S}(\mathcal{K})|=k+1 \Leftrightarrow \mathcal{S}(\mathcal{K})=\mathcal{K}\)
and \(|[\mathcal{K}]| = 1\).
We further call \(\mathbb{S} = \{\mathcal{S}(\mathcal{K}) \mid \mathcal{K}\in B(d,k)\}\) 
the \textbf{complete set of reduced representatives}.
Note that \(\mathbb{S}\cong B(d,k)/\sim\) and thus \(|\mathbb{S}|=|B(d,k)/\sim|\).

To efficiently derandomize our certificates, we can leverage the fact that
we only need a complete set of reduced representatives \(\mathbb{S}\) 
to compute the label probabilities \(p_{v,y}(G)\).
Given \(\mathbb{S}\), we only have to evaluate \(f\) \textit{once} for each reduced representative \(\mathcal{S}(\mathcal{K})\in\mathbb{S}\):
\begin{restatable}{corollary}{propositionLabelProbabilities}\label{prop:labelProbabilities}
    Given the complete set of reduced representatives \(\mathbb{S}\), the label probabilities are:
    \begin{equation*}
        p_{v,y}(G) = \binom{d}{k}^{-1} \sum_{\mathcal{S}\in\mathbb{S}} \mathbb{I}[{f\left(\mathcal{R}_{\mathcal{S}}\right) = y}]\cdot \beta_\mathcal{S}
    \end{equation*}
    where \(\mathbb{I}[{f\left(\mathcal{R}_{\mathcal{S}}\right) = c}]\) 
    indicates whether f classifies the target node \(v\) in subgraph \(\mathcal{R}_\mathcal{S}\) as class c,
    and \(\beta_\mathcal{S}\) is the size of an equivalence class, \(\beta_\mathcal{S} = |[\mathcal{K}]|\).
    We write \(\mathcal{S}\triangleq \mathcal{S}(\mathcal{K})\) and omit \(v\) for conciseness.
\end{restatable}
\begin{proof}
    For all \(\mathcal{K}, \mathcal{K}'\in B(d,k)\) with \(\mathcal{K} \sim \mathcal{K'}\) we have
    \(
    f_v(\mathcal{R}^{v}_\mathcal{K}) 
    = f_v(\mathcal{R}^{v}_\mathcal{K'})
    = f_v(\mathcal{R}^{v}_{\mathcal{S}(\mathcal{K})})
    \)
    as only information from nodes of the reduced representative \(\mathcal{S}(\mathcal{K})\)
    can be passed to the target node (other nodes are disconnected).
    Thus, instead of evaluating \(f_{v}(\mathcal{R}^{v}_{\mathcal{K}}(G))\) for all
    \(\mathcal{K}\in B(d,k)\) we only have to evaluate 
    \(f_{v}(\mathcal{R}^{v}_{\mathcal{S}(\mathcal{K})}(G))\) for each \(\mathcal{S(\mathcal{K})}\in\mathbb{S}\).
    To do so we have to count \({f_{v}(\mathcal{R}^{v}_{\mathcal{S}(\mathcal{K})}(G)) = i}\) 
    exactly \(\beta_\mathcal{S} = |[\mathcal{K}]|\) times.
    Further, as we uniformly sample \(\mathcal{K}\) from \(\mathcal{U}(d,k)\) over \(B(d,k)\), 
    we have to scale the possibilities by \(|B(d,k)|^{-1}\), which corresponds to the inverse binomial coefficient above.
\end{proof}

Hence, we can compute the label probabilities \(p_{v,y}(G)\) exactly for larger receptive fields if
the number of equivalence classes \(|\mathbb{S}|\) is small and
we have an efficient algorithm to compute \(\mathbb{S}\) and \(\beta_\mathcal{S}\).
We propose such algorithm by exploiting the sparsity of graphs as~follows:

We successively enumerate all possible connected subgraphs 
of the receptive field \(\mathcal{R}\) indexed by \(\mathcal{S}\) that contain the target node and at most \(k\) further nodes. Let \(\mathcal{S}\) denote indices of such subgraph of \(\mathcal{R}\)
and \(\mathcal{N}_\mathcal{S}\) the neighborhood of \(\mathcal{S}\) in \(\mathcal{R}\).
If \(\mathcal{S}\) contains \(k + 1\) nodes, then all \(k+1\) nodes will be connected to the target node
and \(\mathcal{S}\) is already a representative with \(\beta_\mathcal{S}=1\). 
If \(\mathcal{S}\) contains less than \(k+1\) nodes, then \(\mathcal{S}\) corresponds to a reduced representative
if we can choose the remaining \(k+1-|\mathcal{S}|\) nodes such that they are disconnected. 
Therefore, the main idea of our algorithm is that 
the size \(\beta_\mathcal{S}\) is just a binomial coefficient:
The number of disconnected nodes is given by \(|V_v| - |\mathcal{N}_\mathcal{S}| - |\mathcal{S}|\),
out of which we have to choose \(k+1-|\mathcal{S}|\) nodes to augment \(\mathcal{S}\) to set of \(k+1\) nodes (where \(V_v\) denote nodes in the receptive field):
\[
    \beta_\mathcal{S} = \binom{|V_v|-|\mathcal{N}_{\mathcal{S}}|-|\mathcal{S}|}{k+1-|\mathcal{S}|}
\]
If \(\beta_\mathcal{S}>0\), there must exist a representative \(\mathcal{K}\) such that 
the reduced representative \(\mathcal{S}(K)\) corresponds to \(\mathcal{S}\),
that is \(\mathcal{S}=\mathcal{S}(\mathcal{K})\) (compare (4) in \autoref{fig:eqc} for an example).
Finally, our algorithm enumerates all possible \(\mathcal{S}\) 
by recursively augmenting \(\mathcal{S}\) with nodes from the neighborhood of \(\mathcal{S}\) (compare \autoref{algo1}). This way, we exploit the sparsity of graphs to find all reduced representatives \(\mathbb{S}\)
that avoid disconnected nodes.
\clearpage

\begin{algorithm}
  \DontPrintSemicolon
    \caption{Compute complete set of reduced representatives \(\mathbb{S}\) and equivalence class sizes \(\beta_\mathcal{S}\)}\label{algo1}
    \SetKwFunction{FEQG}{EQCGeneration}
    \SetKwFunction{Fupdate}{updateDistances}
    \SetKwFunction{Fbinom}{binom}

    \KwIn{
     Index 0 of target node \(v\),
     Receptive field \(\mathcal{R}^{v}=(V_v,E_v)\),
     Retention constant \(k\)
     }
     \(\mathcal{S} \gets \{0\}\)\;
     \KwOut{\FEQG{\(\mathcal{S}\), \(V_v\), \(E_v\), \(k\)}}\;
  \SetKwProg{Fn}{Function}{:}{}
  \Fn{\FEQG{\(\mathcal{S}\), \(V_v\), \(E_v\), \(k\)}}{
      \(R \gets \{\}\)\;
      \If{\(|\mathcal{S}| = k+1\)}{
          \KwRet \(\{(\mathcal{S},1)\}\)\;
      }
      \(\mathcal{N}_{\mathcal{S}} \gets \{w\in V_v\setminus\mathcal{S} \mid \exists u\in \mathcal{S} : (w,u)\in E_v\}\) \tcp*[r]{\(\mathcal{O}(|V_v|)\)}
      \(\beta_\mathcal{S} \gets\) \Fbinom{\(|V_v|-|\mathcal{N}_{\mathcal{S}}|-|\mathcal{S}|,k+1-|\mathcal{S}|\)}\;
      \If{\(\beta_\mathcal{S}>0\)}{
          \(R\gets \{(\mathcal{S},\beta_\mathcal{S})\}\)\;
      }
      \For(\tcp*[f]{\(\mathcal{O}(|V_v|)\)}){\(w\in \mathcal{N}_{\mathcal{S}}\)}{
          \(R\gets R\, \cup\,\)\FEQG{\(\mathcal{S}\cup \{w\}\), \(V_v\), \(E_v\), \(k\)}\;
      }
      \KwRet{R}
  }
\end{algorithm}
Note that in \autoref{algo1}, \(V_v\) denotes nodes in the receptive field of classifier \(f\) with respect to target node \(v\), and \(E_v\) the edges in the receptive field.

\begin{restatable}[Correctness of \autoref{algo1}]{lemma}{lemmaCorrectnessAlgorithm}\label{lemma:correctness}
    Let \(\mathcal{S}\) with \(0\in\mathcal{S}\subseteq V_v\) 
    be a set of at most \(k+1\) nodes \(1\leq |\mathcal{S}|\leq k+1\) such that 
    all nodes indexed by \(\mathcal{S}\) are connected to the target node in \(\mathcal{R}\).
    We denote the neighbors of \(\mathcal{S}\) in \(\mathcal{R}\) as
    \(\mathcal{N}_{\mathcal{S}} \triangleq \{w\in V_v\setminus\mathcal{S} \mid \exists u\in \mathcal{S} : (w,u)\in E_v\}\).
    When we define the following binomial coefficient as
    \[
        \beta_\mathcal{S} \triangleq \binom{|V_v|-|\mathcal{N}_{\mathcal{S}}|-|\mathcal{S}|}{k+1-|\mathcal{S}|} \in\mathbb{N}.
    \]
    then there exists a representative \(\mathcal{K} \in B(d,k)\) 
    such that \(\mathcal{S}\) is a reduced representative for the equivalence class \([\mathcal{K}]\) if \(\beta_\mathcal{S}>0\).
    Then we have \(\beta_\mathcal{S} = |[\mathcal{K}]|\).
\end{restatable}
\begin{proof}
    First note that for a given set \(\mathcal{S}\) as defined above we can partition \(V_v\) into three disjoint sets
    \(V_v=\mathcal{S} \uplus \mathcal{N}_{\mathcal{S}} \uplus \mathcal{N}_r\)
    with \(\mathcal{S}\) and \(\mathcal{N}_\mathcal{S}\) defined as above,
    and the disconnected nodes \(\mathcal{N}_r \triangleq V_v\setminus(\mathcal{S}\cup \mathcal{N}_\mathcal{S})\).
    We thus have \(|\mathcal{N}_r| = |V_v|-|\mathcal{N}_{\mathcal{S}}|-|\mathcal{S}|\).
    Now we distinguish the following cases:

    \underline{Case 1}: \(|\mathcal{S}|=k+1\)

    We have \(|V_v|-|\mathcal{N}_{\mathcal{S}}|-|\mathcal{S}| \in\mathbb{N}_0\) and \( \beta_\mathcal{S} = 1 > 0 \).
    Thus for \(|\mathcal{S}|=k+1\) the condition is trivially fulfilled and
    we have that \(\mathcal{K} \triangleq \mathcal{S}\) is already a representative
    with \(|[\mathcal{K}]| = 1\) as discussed before.
    Note that this does not mean that all sets with \(k+1\) nodes are representatives,
    as we still have the connectivity constraint for nodes in \(\mathcal{S}\).

    \underline{Case 2}: \(|\mathcal{S}|<k+1\) 

    We have
    \(  
        \beta_\mathcal{S} > 0 
        \Leftrightarrow
        |V_v|-|\mathcal{N}_{\mathcal{S}}| - |\mathcal{S}| \geq k+1-|\mathcal{S}|
        \Leftrightarrow
        |\mathcal{N}_r| \geq k+1-|\mathcal{S}|
    \)
    where the latter means that we can choose the remaining \(k+1-|\mathcal{S}|\)
    nodes from \(\mathcal{N}_r\) to augment 
    \(\mathcal{S}\) to representative \(\mathcal{K}\) of the equivalence class \([\mathcal{K}]\)
    since then \(|\mathcal{K}| = |\mathcal{S}| + k + 1 - |\mathcal{S}| = k+1\).
    The corresponding size \(|[\mathcal{K}]|\) is given by \(\beta_\mathcal{S}\).
\end{proof} 

Finally, note that the equivalence classes and the algorithm are independent of the classifier \(f\).

\clearpage
\textbf{Discussion.}
In the worst case, we have \(|\mathbb{S}| = |B(d,k)|=\binom{d}{k}\),
but we enumerate \(\sum_{i=0}^k\binom{d}{i} \geq \binom{d}{k} \) possibilities,
as there are \(\sum_{i=0}^k\binom{d}{i}\) candidates for reduced representatives in a fully connected graph.
Therefore, in the worst case of fully connected graphs, directly enumerating all \(\binom{d}{k}\) possibilities
would be faster. 
In practice, however, we rather observe sparse graphs with \(|\mathbb{S}| \ll |B(d,k)|\).
The more sparse the receptive field, the less equivalence classes exist and the larger each equivalence class.
Thus we exploit the sparsity of graphs to efficiently compute \(\mathbb{S}\) 
and the corresponding sizes \(|[\mathcal{K}]|\) for all equivalence classes~\([\mathcal{K}]\).

Moreover, as our algorithm recursively enumerates all possible pairs \((\mathcal{S}, \beta_\mathcal{S})\),
we can determine a stopping criterion at which we back off to Monte-Carlo sampling for estimating the label probabilities. 
To this end, if \(R\) denotes the current set of \((\mathcal{S}, \beta_\mathcal{S})\) pairs with \(\beta_\mathcal{S}>0\),
we know that \(|R|\) is a lower bound on the number of equivalence classes, \(|R| \leq |\mathbb{S}|\).
By summing up \(\beta_\mathcal{S}\) for all \((\mathcal{S}, \beta_\mathcal{S})\in R\)
we can determine the percentage of \(|B(d,k)|\) that we already cover with \(R\):
\[ 
    \sum_{(\mathcal{S},\beta_\mathcal{S})\in R} \beta_\mathcal{S} 
    \leq \sum_{\mathcal{S}(\mathcal{K})\in\mathbb{S}} |[\mathcal{K}]|
    = \binom{d}{k} 
    = |B(d,k)|
\]
This allows us to use the condition \(\sum_{(\mathcal{S}, \beta_\mathcal{S})\in R} \beta_\mathcal{S} > \tau'\) with
threshold \(\tau'\in\mathbb{N}\) as a stopping criterion.
Using thresholds this way,
our algorithm will always find more solutions in \(\mathbb{S}\) given more time via larger thresholds.
Note that we use \( \binom{d}{k} > \tau\) in practice,
since the binomial coefficient provides a fast upper bound for the number of equivalence classes \(|\mathbb{S}|\).

\subsection{Evaluating Message-passing-aware Derandomization}
\begin{table}[h]
  \footnotesize%
  \renewcommand{\arraystretch}{1.00}%
  \setlength{\tabcolsep}{2pt}%
  \caption{Smoothed classifier results for GCN trained on Cora-ML for different relative retention constants.
          Der.: Ratio of nodes with derandomized certificates.
          Eq.: Mean of unique receptive fields over all derandomized certificates.
          Acc.: Clean accuracy.
  } \label{table:basecertresults}
  \centering 
  \begin{tabular}{@{}ccccccccccccc@{}}
      \toprule
      & \multicolumn{4}{c}{\textbf{GCN on Cora-ML}}
      & \multicolumn{4}{c}{\textbf{GCN on Citeseer}}
      & \multicolumn{4}{c}{\textbf{GCN on PubMed}} \\
      \cmidrule(lr){2-5}
      \cmidrule(lr){6-9}
      \cmidrule(lr){10-13}
      \(k_{rel}\) & Der. & Eq. & Abstained & Acc.
                  & Der. & Eq. & Abstained & Acc.
                  & Der. & Eq. & Abstained & Acc. \\
      \midrule
      0.01 & 0.87 & 0.22 & 6.27e-04 & 0.73 & 1.00 & 0.41 & 0.00e+00 & 0.65 & 0.94 & 0.15 & 0.00e+00 & 0.73 \\
      0.03 & 0.72 & 0.23 & 5.69e-04 & 0.73 & 0.94 & 0.42 & 0.00e+00 & 0.66 & 0.81 & 0.16 & 1.56e-03 & 0.73 \\
      0.10 & 0.50 & 0.28 & 5.02e-03 & 0.74 & 0.87 & 0.42 & 1.63e-03 & 0.65 & 0.61 & 0.19 & 4.24e-03 & 0.74 \\
      0.30 & 0.31 & 0.46 & 1.42e-02 & 0.80 & 0.73 & 0.53 & 7.61e-03 & 0.68 & 0.37 & 0.38 & 6.23e-03 & 0.77 \\ 
      \bottomrule
  \end{tabular}
\end{table}
\textbf{Relative retention constant.} 
Consider a small retention constant \(k=1\) for a node \(v\) with \(deg(v) < d_v-deg(v)\), where \(d_v\) denotes the receptive field size (excluding the target node).
Then the probability for selecting a direct neighbor of \(v\) is low
and the prediction of the smoothed classifier is merely based on the target node \(v\) itself, 
which amounts to traditional i.i.d. prediction.
Thus, for non-trivial robustness guarantees we use 
retention constants \(k\) that are relative to the receptive field size:
Given a fixed relative retention constant \(k_{rel}\in [0,1]\),
our smoothed classifier keeps \(k = \lceil d_v\cdot k_{rel} \rceil \in \mathbb{N}\) nodes 
in the receptive field~\(\mathcal{R}\).%
\footnote{As a disadvantage of this method, we have to process all receptive fields separately.} The ceiling operation ensures that we keep at least one additional~node.

\textbf{Derandomization results.}
Our certificates are deterministic for small receptive fields, and probabilistic for large receptive fields:
we derandomize certificates if \(\binom{d}{k}\) is smaller than a threshold \(\tau\).
If the number of possibilities to choose \(k\) out of \(d\) nodes is small,
we can enumerate all possibilities and use \(f\) to predict the class of \(v\) for all possibilities.
In our experiments we set \(\tau = 100{,}000\).
There are more possibilities to sample \(k\) out of \(d\) nodes for larger \(k_{rel}\) 
and thus the ratio of deterministic certificates decreases (compare \autoref{table:basecertresults}).
For example, we can derandomize around 50\% of the certificates for Cora-ML given \(k_{rel} = 0.1\).
We further derandomize more certificates for Citeseer than for Cora-ML,
which can be explained by the fact that two-layer GNNs have larger receptive fields on Cora-ML.
Note that the average degree in Cora-ML is \(6\), in Citeseer \(3\) and PubMed~\(4\).
Due to the derandomization we also hardly observe that the smoothed classifier abstains.

As discussed above, we avoid evaluating the base classifier \(f\) for equivalent receptive fields. To represent the computations we avoid on average,
we compute the mean of unique receptive fields 
\(|\mathbb{S}|/|B(d,k)|\) for all derandomized certificates.
For example, out of all derandomized certificates for \(k_{rel}=0.1\) on Cora-ML, 
we only have to evaluate 28\% of all possibilities on average.